%% file: agglo_clust_review.tex
\pdfoutput=1
\documentclass[10pt,twocolumn,letterpaper]{article}

\usepackage{cvpr}              %

\makeatletter
\@namedef{ver@everyshi.sty}{}
\makeatother

\usepackage{graphicx}
\usepackage{booktabs}

\usepackage{amsmath}
\usepackage{amssymb}

\usepackage{afterpage}
\usepackage{tikz}
\usetikzlibrary{matrix,positioning,calc}
\tikzset{line/.style ={draw, rounded corners=2pt, line width=1pt}}

\newcommand\tikzmark[1]{%
\tikz[remember picture]  \node[inner sep=0,outer sep=0] (#1){};%
}

\usepackage{amsthm}
\usepackage{algorithm}%
\PassOptionsToPackage{noend}{algpseudocode}%
\usepackage{algpseudocode}%
\usepackage{marvosym}

\makeatletter

\renewcommand{\ALG@beginalgorithmic}{\small}
\makeatother

\usepackage{etoolbox}

\usepackage{tabularx}
\usepackage{lipsum}
\usepackage{makecell}

\usepackage{multirow}
\usepackage[labelformat=simple]{subcaption}

\usepackage{booktabs}
\usepackage{rotating}
\newcolumntype{?}{!{\vrule width 0.3em}}

\algrenewcommand\algorithmicindent{0.8em}
\newcommand*{\cost}{w}%
\newcommand*{\interact}{\mathcal{W}}
\newcommand*{\NBE}{\Gamma} %
\newcommand*{\nBE}{\gamma}
\newcommand*{\algname}{GASP}
\newcommand*{\treeHeight}{\interact{}_{T}}
\newcolumntype{M}[1]{>{\centering\arraybackslash}m{#1}}
\newcolumntype{R}[1]{>{\raggedleft\arraybackslash}m{#1}} 
\newcolumntype{L}[1]{>{\raggedright\arraybackslash}m{#1}} 
 \newcolumntype{?}{!{\vrule width 0.3em}}

\DeclareMathOperator*{\argmax}{arg\,max}

\usepackage{mathrsfs}

\usepackage[shortlabels]{enumitem}

\usepackage{thmtools}
\usepackage{thm-restate}
\newtheorem{theorem}{Theorem}[section]
\newtheorem{prop}{Proposition}[section]

\newtheorem{lemma}[theorem]{Lemma}

\theoremstyle{definition}

\theoremstyle{remark}

\newif\ifappendix
\appendixtrue

\usepackage[pagebackref,breaklinks,colorlinks]{hyperref}

\usepackage[capitalize]{cleveref}
\crefname{section}{Sec.}{Secs.}
\Crefname{section}{Section}{Sections}
\Crefname{table}{Table}{Tables}
\crefname{table}{Tab.}{Tabs.}

\def\cvprPaperID{2544} %
\def\confName{CVPR}
\def\confYear{2022}

\begin{document}

\title{GASP, a generalized framework for agglomerative clustering of signed graphs and its application to Instance Segmentation} %

\author{Alberto Bailoni \thanks{Heidelberg Collaboratory for Image Processing (HCI/IWR), Heidelberg, Germany. {\tt \{name.surname\}@iwr.uni-heidelberg.de}}~~\thanks{EMBL, Heidelberg, Germany. {\tt \{name.surname\}@embl.de}}
\and Constantin Pape \thanks{Institute for Computer Science, University G\"ottingen, Germany.
{\tt constantin.pape@informatik.uni-goettingen.de}}~~\footnotemark[2]~~\footnotemark[1]
\and Nathan H\"utsch \footnotemark[1]
\and Steffen Wolf \thanks{MRC Laboratory of Molecular Biology, Cambridge, UK.  {\tt swolf@mrc-lmb.cam.ac.uk}}
\and Thorsten Beier \thanks{{\tt derthorstenbeier@gmail.com}}
\and Anna Kreshuk \footnotemark[2]
\and Fred A. Hamprecht \footnotemark[1]~~\thanks{Corresponding author}
}

\maketitle

\input{chapters/0_abstract.tex}

\input{chapters/1_intro.tex}
\input{chapters/2_rel_work}

\input{chapters/3_algorithm}

\input{chapters/5_experiments_intro}

\input{chapters/5_experiments_compare_rules}

\input{chapters/6_conclusions}

{\small
\bibliographystyle{ieee_fullname}
\bibliography{agglo_clust_review}
}

\ifappendix  
\clearpage
\input{chapters/supplementary}

\else  
\fi

\end{document}

%% file: chapters/0_abstract.tex
\begin{abstract}
We propose a theoretical framework that generalizes simple and fast algorithms for hierarchical agglomerative clustering to weighted graphs with both attractive and repulsive interactions between the nodes. This framework defines GASP, a Generalized Algorithm for Signed graph Partitioning\footnote{Code available at: \url{https://github.com/abailoni/GASP}}, and allows us to explore many combinations of different linkage criteria and cannot-link constraints. 
We prove the equivalence of existing clustering methods to some of those combinations and introduce new algorithms for combinations that have not been studied before. 
We study both theoretical and empirical properties of these combinations and prove that some of these define an ultrametric on the graph.
We conduct a systematic comparison of various instantiations of GASP on a large variety of both synthetic and existing signed clustering problems, in terms of accuracy but also efficiency and robustness to noise. 
Lastly, we show that some of the algorithms included in our framework, when combined with the predictions from a CNN model, result in a simple bottom-up instance segmentation pipeline.
Going all the way from pixels to final segments with a simple procedure, we achieve state-of-the-art accuracy on the CREMI 2016 EM segmentation benchmark without requiring domain-specific superpixels.
\end{abstract}

%% file: chapters/1_intro.tex
\begin{figure*}[t]
\centering
\includegraphics[width=0.92\textwidth]{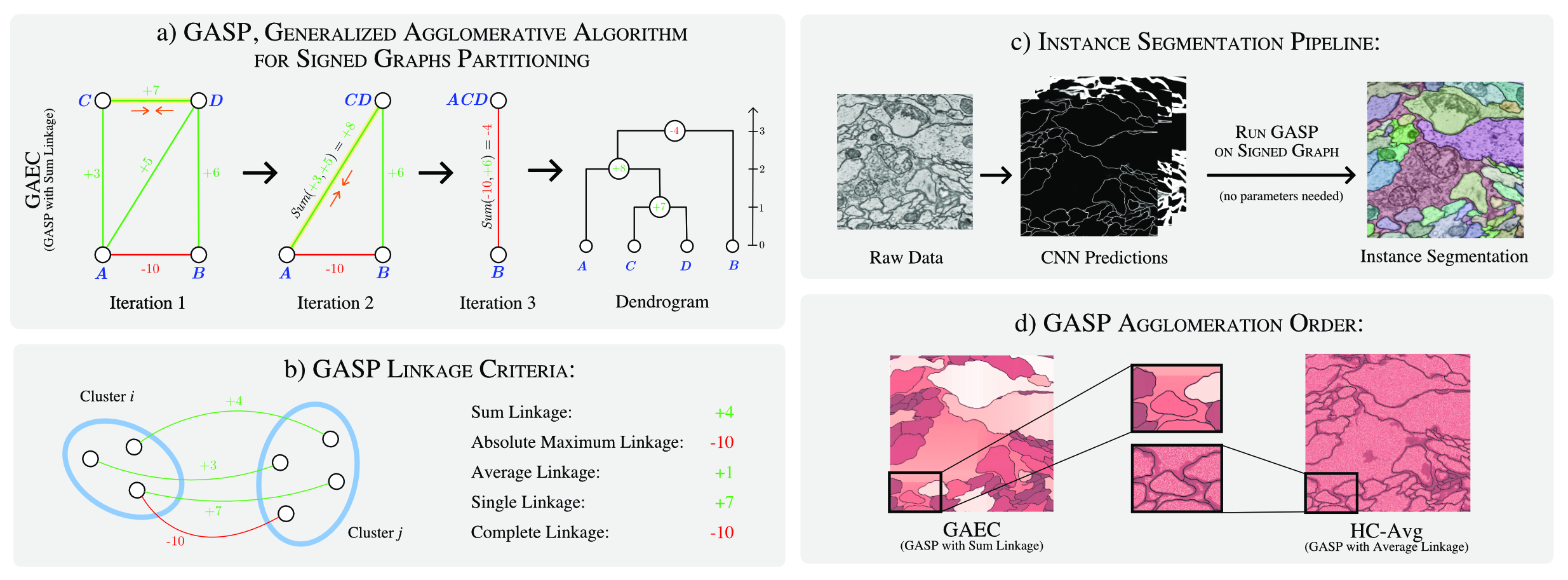} %
\caption{\textbf{(a)} Some iterations of \algname{} on a graph with attractive (green) and repulsive (red) interactions. At each iteration, the yellow edge with highest weight is contracted (example with sum linkage criterion is shown). \textbf{(b)} Linkage criteria used in this paper demonstrated on two small clusters (see definitions in Table~\ref{tab:linkage-criteria} below).  \textbf{(c)} Application of \algname{} to instance segmentation: we show raw data from the CREMI neuron-segmentation challenge and some predictions of our CNN model, where white pixels represent boundary evidence. \textbf{(d)} 
Seemingly similar linkage criteria can result in very different clustering dynamics, as shown in this example: color coded sequence of merges from early (white) via late (brown) to never (black).
\label{fig:intro_figure}}
\end{figure*}

\begin{table*}[t]
    \centering
    \footnotesize
    \begin{subtable}[t!]{\textwidth}\centering
        \begin{tabular}{l |c  c  c  c  c}
        \multicolumn{1}{c|}{\multirow{2}{*}[-1.5em]{{\Large GASP}}} 
        & \thead{Sum\\Linkage} & \thead{Absolute Maximum\\Linkage} & \thead{Average\\Linkage} & \thead{Single\\Linkage} & \thead{Complete\\Linkage} \\
 & $\displaystyle \sum_{e\in E_{ij}} \cost_e$  & $\displaystyle \cost_e$ with $\displaystyle e = \argmax_{t\in E_{ij}} |\cost_t|$ & $\displaystyle \sum_{e\in E_{ij}} \cost_e \bigg/ \big|E_{ij}\big| $ &  $\displaystyle \max_{e\in E_{ij}} \cost_e$ & $\displaystyle \min_{e\in E_{ij}} \cost_e$ \\ \midrule

            \thead[r]{Unsigned graphs} & \thead{-} &\tikzmark{a} \thead{\textbf{HC-Single}} &\tikzmark{g} \thead{\textbf{HC-Avg}} &\thead{\textbf{HC-Single}}\tikzmark{z} &\thead{\textbf{HC-Complete}} \tikzmark{b} \\
            \thead[r]{Signed graphs} & \thead{GAEC \cite{keuper2015efficient}} &  \thead{\textbf{Mutex Watershed} \cite{wolf2018mutex}}& \tikzmark{f} \thead{\textbf{HC-Avg}} &\thead{\textbf{HC-Single}} \tikzmark{e} &\thead{\textbf{HC-Complete}}\tikzmark{w} \\
            \thead[r]{Signed graphs + cannot-link constraints} & \thead{\colorbox{yellow}{HCC-Sum}} %
            & \thead{\textbf{Mutex Watershed} \cite{wolf2018mutex}}& \thead{\colorbox{yellow}{HCC-Avg}} &  \thead{\colorbox{yellow}{HCC-Single}} \tikzmark{d} &   \thead{\textbf{HC-Complete}} \tikzmark{c} \\

        \end{tabular}
    \end{subtable} 
    \caption{Conceptual contribution: Properties of clustering algorithms included in the proposed \algname{} framework, given a linkage criterion, a type of graph (signed or unsigned) and the optional use of cannot-link constraints. New constrained hierarchical clustering algorithms (HCC) proposed in this paper are highlighted in yellow. For algorithms typeset in bold font we prove that they define an ultrametric on the graph (Eq.~\ref{eq:UM_def}). For algorithms in the green box we show that they are weight-shift invariant (Prop.~\ref{prop:weight_shift_invariant}). 
    Notation: 
    $E_{ij}$ denotes the set of edges connecting two clusters $S_i, S_j \subseteq V$. } 
    \label{tab:linkage-criteria}
\end{table*}

\section{Introduction}
In computer vision, the partitioning of weighted graphs has been successfully applied to tasks as diverse as image segmentation, object tracking and pose estimation. 
Most graph clustering methods work with positive edge weights only, which can be interpreted as similarities or distances between the nodes. These methods require users to specify the desired numbers of clusters (as in spectral clustering) or a termination criterion (e.g.\ in iterated normalized cuts) or even to add a seed for each object  (e.g.\ seeded watershed or random walker).  

Other graph clustering methods work with so-called \emph{signed graphs}, which feature both positive and negative edge weights corresponding to attraction and repulsion between nodes. The advantage of signed graphs over unsigned graphs is that balancing attraction and repulsion allows us to obtain a clustering without defining additional parameters. A canonical formulation of the signed graph partitioning problem is the \emph{multicut} or \emph{correlation clustering} problem \cite{kappes2011globally,chopra1991multiway}. This problem is NP-hard, though many approximate solvers have been proposed \cite{lange2018combinatorial,pape2017solving,beier2016efficient,yarkony2012fast} together with greedy agglomerative clustering algorithms \cite{keuper2015efficient,levinkov2017comparative,wolf2018mutex,kardoostsolving}. 
Agglomerative clustering algorithms for signed graphs have clear advantages: they are parameter-free and efficient. Despite the fact that a variety of these algorithms exist, no overarching study has so far been conducted to compare their robustness and efficiency or to provide guidelines for matching an algorithm to the partitioning problem at hand.

Our first contribution is a simple theoretical framework that generalizes over agglomerative algorithms for signed graphs by linking them to hierarchical clustering (HC) on unsigned graphs (Section \ref{sec:algorithm}). This framework defines an underlying basic algorithm and allows us to explore its combinations with different linkage criteria and \emph{cannot-link constraints} (see Fig.~\hyperref[fig:intro_figure]{\ref*{fig:intro_figure}a}, \hyperref[fig:intro_figure]{\ref*{fig:intro_figure}b}, and Table~\ref{tab:linkage-criteria}). 
As second contribution, in Section \ref{sec:alg_update_rules}, we formally prove that some of the combinations correspond to existing clustering algorithms, and introduce new algorithms for combinations which have not been explored before. By analyzing their theoretical properties, we also show that some of them define an ultrametric on the graph (see Table~\ref{tab:linkage-criteria}).

Third, we evaluate the algorithms on a large variety of both existing and synthetically generated signed graph clustering problems (Section \ref{sec:neuro_segm_exp}). 
Fourth and finally, we also test the algorithms on \emph{instance segmentation} -- a computer vision task consisting of assigning each pixel of an image to an object instance -- by partitioning graphs whose edge weights are estimated by a CNN (see Fig.~\hyperref[fig:intro_figure]{\ref*{fig:intro_figure}c} and Section \ref{sec:experiments_discussion}).
Our experiments show that the choice of linkage criterion markedly influences how clusters are grown by the agglomerative algorithms (Fig.~\hyperref[fig:intro_figure]{\ref*{fig:intro_figure}d}), making some linkage methods more suited for certain types of clustering problems.
We benchmark the clustering algorithms by focusing on their efficiency, robustness and tendency to over- or under-cluster. 
On instance segmentation, we show that the agglomerative algorithms outperform recently proposed spectral clustering methods, and that average-linkage based agglomerative algorithms achieve state of the art results on the CREMI 2016 challenge for neuron segmentation of 3D electron microscopy image volumes of brain tissue. 

%% file: chapters/2_rel_work.tex
\begin{tikzpicture}[remember picture,overlay]
\draw [green,line]($(a)+(0,2ex)$)--($(b)+(0.9ex,2ex)$)--($(c)+(0.9ex,-1.5ex)$) -- ($(d)+(1.4ex,-1.5ex)$) -- ($(e)+(1.4ex,-1.2ex)$) -- ($(f)+(-0.3ex,-1.2ex)$)--($(g)+(-0.3ex,-1.2ex)$)--($(a)+(0,-1.2ex)$)--cycle;
\end{tikzpicture}
\section{Related work} \label{sec:related_work}

\textbf{Proposal-free instance segmentation methods} adopt a bottom-up approach by directly grouping pixels into instances. In the last years, there has been a growing interest in such  methods that do not involve object detection because, in certain types of data, object instances cannot be approximated by bounding boxes \cite{kirillov2017instancecut,bai2017deep}. 
Some use metric learning to predict high-dimensional associative pixel embeddings that map pixels of the same instance close to each other \cite{lee2019learning,fathi2017semantic,newell2017associative,de2017semantic}
and then retrieve final instances by applying a clustering algorithm \cite{kong2018recurrentPix}.
Other recent methods let the model predict the relative coordinates of the instance center \cite{neven2019instance,cheng2019panopticdeeplab} or, given a pixel $(x,y)$, they train a model to generate the mask of the instance located at $(x,y)$ \cite{sofiiuk2019adaptis}. 

\textbf{Edge detection} also experienced recent progress thanks to deep learning, both on natural images \cite{Gao_2019_ICCV,liu2018affinity,xie2015holistically,kokkinos2015pushing} and biological data \cite{lee2017superhuman,schmidt2018cell,meirovitch2016multi,ciresan2012deep}. In neuron segmentation for connectomics, a field of neuroscience we also address in our experiments, boundaries are converted to final instances with subsequent postprocessing and superpixel-merging:
some use a combinatorial framework \cite{beier2017multicut}, others use loopy graphs \cite{kaynig2015large,krasowski2015improving} or trees \cite{meirovitch2016multi,liu2016sshmt,liu2014modular,funke2015learning,uzunbas2016efficient} to represent the region merging hierarchy. Flood-filling networks \cite{januszewski2018high} and MaskExtend \cite{meirovitch2016multi} used a CNN to iteratively grow one region/neuron at the time.
A structured learning approach was also proposed in \cite{funke2018large,turaga2009maximin}.

\textbf{Agglomerative graph clustering} has often been applied to instance segmentation \cite{arbelaez2011contour,ren2013image,liu2016image,salembier2000binary} because of its efficiency as compared to other divisive approaches like graph cuts. 
Novel termination criteria and merging strategies have often been proposed: the agglomeration in \cite{malmberg2011generalized} deploys fixed sets of merge constraints; 
the popular graph-based method \cite{felzenszwalb2004efficient} stops the agglomeration when the merge costs exceed a measure of quality for the current clusters. 
The optimization approach in \cite{kiran2014global} performs greedy merge decisions that minimize a certain energy, while other pipelines use classical linkage criteria, e.g.~average linkage \cite{liu2018affinity,lee2017superhuman}, median \cite{funke2018large} or a linkage learned by a random forest classifier \cite{nunez2013machine,knowles2016rhoananet}.

\textbf{Clustering of signed graphs} has the goal of partitioning a graph with both attractive and repulsive cues. Finding an optimally balanced partitioning has a long history in combinatorial optimization \cite{grotschel1989cutting,grotschel1990facets,chopra1993partition}. %
NP-hardness of the \emph{correlation clustering} problem was shown in \cite{bansal2004correlation}, while the connection with graph multicuts was made by \cite{demaine2006correlation}. Modern integer linear programming solvers can tackle problems of considerable size \cite{andres2012globally}, but accurate approximations \cite{pape2017solving,beier2016efficient,yarkony2012fast}, greedy agglomerative algorithms \cite{levinkov2017comparative,wolf2019mutex,keuper2015efficient,kardoostsolving} and persistence criteria \cite{lange2018partial,lange2018combinatorial} have been proposed for even larger graphs. 
Another line of research is given by spectral clustering methods that, on the other hand, require the user to specify the number of clusters in advance. Recently, some of these methods have been generalized to graphs with signed weights \cite{Cucuringu2019SPONGEAG,chiang2012scalable,kunegis2010spectral}, whereas others let the user specify must-link and cannot-link constraints between clusters \cite{rangapuram2012constrained,wang2014constrained,cucuringu2016simple}.

This work reformulates the clustering algorithms of \cite{levinkov2017comparative,wolf2018mutex,keuper2015efficient} in a generalized framework and adopts ideas from the proposal-free instance segmentation methods \cite{liu2018affinity,wolf2018mutex,lee2017superhuman} to predict edge weights of a graph.

%% file: chapters/3_algorithm.tex
\section{Generalized framework for agglomerative clustering of signed graphs} \label{sec:general_framework}

\subsection{Notation} \label{sec:notation}

\textbf{Graph formalism} -- We consider an undirected simple edge-weighted graph $\mathcal{G}(V,E,w^+, w^-)$ with both attractive and repulsive edge attributes.
The weight function $w^+: E \rightarrow \mathbb{R}^+$ associates to every edge a positive scalar attribute $w_e^+\in \mathbb{R}^+$ representing a merge affinity or a similarity measure.
On the other hand, $w^-: E \rightarrow \mathbb{R}^+$ associates to each edge a split tendency $w_e^- \in \mathbb{R}^+$.
Graphs of the type $\mathcal{G}(V,E,w^+, w^-)$ are often defined as \emph{signed graphs} $\mathcal{G}(V,E,\cost)$, featuring positive and negative edge weights $\cost_e\in \mathbb{R}$. Following the theoretical considerations in \cite{lange2018partial}, we define signed weights as ${\cost_e = w_e^+ - w_e^-}$. 

\textbf{Multicut objective} -- We call the set $\Pi=\{S_1,\ldots,S_K\}$ a \emph{clustering} or \emph{partitioning} if $V = \cup_{S\in\Pi} S $, $\,S \cap S' = \emptyset$ for different clusters $S, S'$ and every cluster $S \in \Pi$ induces a connected subgraph of $\mathcal{G}$. 
For any clustering $\Pi$ of $\mathcal{G}$, we denote as $E^0_\Pi= \{ e_{uv} \in E \,|\, \exists S \in \Pi : u,v \in S \}$ the set of edges linking nodes in the same cluster. Its complementary set $E_\Pi^1= E \setminus E^0_\Pi$ of edges linking nodes belonging to distinct clusters, is known as the \emph{multicut} of $\mathcal{G}$ associated to clustering $\Pi$. The instance of the NP-hard \emph{minimum cost multicut problem} w.r.t. $\mathcal{G}(V,E,w_e)$ is the task of finding a clustering that optimally balances the attraction and repulsion in the graph and is given by the following binary integer program:
\begin{equation}\label{eq:MC_objective}
 \min_\Pi \sum_{e\in E} \cost_e x_e^\Pi,  \qquad \text{where} \quad x^\Pi_e = 
 \begin{cases} 
 1 & \text{if } e\in E^1_\Pi \\
 0 & \text{otherwise}.
 \end{cases}
\end{equation}

\textbf{Linkage criteria and hierarchical trees} -- 
Let the interaction $\interact(S \cup S')\in\mathbb{R}$ between two clusters $S,S'$ be defined as a function, named \emph{linkage criterion}, depending on the weights of \emph{all} edges connecting clusters $S$ and $S'$.
The linkage criteria tested in this article are listed and defined in Table \ref{tab:linkage-criteria}.
A \emph{dendrogram} $T$ is a rooted binary tree\footnote{In general, one could look at trees that are not binary. However, the algorithms discussed in this paper always generate binary hierarchical trees, so nothing would be gained by this generalization.} representing the merging order of an agglomerative algorithm, such that the leaves of the tree are in one-to-one correspondence with $V$ and each node of the tree represents a merge between two clusters. 
Let $T_{\mathrm{R}},T_{\mathrm{L}}\subset T$ denote the subtrees rooted at the two children of the root node in $T$.
For any two leaves $u,v \in V$, let $T[u \vee v]$ be the subtree rooted at the least common ancestor $(u \vee v)\in T$ of nodes $u$ and $v$ (furthest from the root), and let \texttt{leaves}$(T[u \vee v])\subseteq V$ be the set of leaves of this subtree. 
Given an agglomerative algorithm with merging tree $T$, let $h_T:V \times V \rightarrow \mathbb{N}$ denote the \emph{dendrogram-height} of each $(u\vee v)\in T$, which is defined as the iteration number at which nodes $u,v\in V$ were merged by the algorithm (see example in Fig.~\ref{fig:intro_figure}a). We also define $\treeHeight(u,v)$ as the signed interaction $\interact{}(S \cup S')$ between the two clusters $S,S'$ that were merged at iteration $h_T(u, v)$: 
\begin{equation}\label{eq:def_dendr_interact}
\treeHeight(u,v) \equiv \interact{} \big( \text{\texttt{leaves}}(T_{\mathrm{R}}[u \vee v])  \cup  \text{\texttt{leaves}}(T_{\mathrm{L}}[u \vee v]) \big)
\end{equation}

\subsection{The \algname{} algorithm} \label{sec:algorithm} 

Our main contribution is a generalized agglomerative algorithm for signed graph partitioning (GASP) that generalizes hierarchical clustering (HC) to signed graphs. 
The framework, defined in the following, encompasses several known and new agglomerative algorithms on display in Table \ref{tab:linkage-criteria}, which are differentiated by the linkage criterion employed, similarly to HC.

In Algorithm \ref{main_alg}, we provide simplified pseudo-code for the proposed \algname{} algorithm. \algname{} implements a bottom-up approach that starts by assigning each node to its own cluster and then iteratively merges pairs of adjacent clusters. The algorithm proceeds in three phases. 

In phase one, \algname{} selects the pair of clusters with the highest absolute interaction $|\interact(S  \cup  S')|$, so that the most attractive and the most repulsive pairs are analyzed first. If the interaction is repulsive and the algorithm option \emph{addCannotLinkConstraints} is \texttt{True}, then the two clusters are constrained so that their members can never merge in subsequent steps of phase one. If the interaction is attractive, then the clusters are merged, provided that they were not previously constrained. 
After each merge, the interaction between the merged cluster and its neighbors is updated according to one of the linkage criteria $\interact(S \cup  S')$ listed in Table \ref{tab:linkage-criteria}. Phase one terminates when all the remaining clusters are either constrained or share repulsive interactions. Note that, on unsigned graphs, in phase one all nodes are merged into a single cluster and \algname{} is then equivalent to a standard hierarchical clustering algorithm.

Phase two: Now that the clusters have grown in size, the algorithm removes the constraints previously introduced in phase one and merges all the clusters that still share an attractive interaction, merging the most attractive one first\footnote{Note that in the version of \algname{} without \emph{cannotLinkConstraints}, nothing happens in phase two because all remaining interactions are repulsive.}. The final clustering $\Pi^*$ returned by \algname{} is found at the end of phase two and it is then composed of clusters sharing only mutual repulsive interactions. 

Finally, in phase three, the algorithm keeps merging all clusters until only a single one is left and then returns the hierarchical tree $T^*$ representing the full sequence of merging steps.
The algorithm was implemented using a standard HC implementation with computational complexity $\mathcal{O}(N^2 \log N)$ (details left in Appendix 
\ifappendix  
\ref{sec:detailed_impl}). 
\else  
A6.1). 
\fi

\begin{algorithm}[t]
\footnotesize
  \begin{flushleft}
  \footnotesize
  \caption{\algname{}}
   \hspace*{\algorithmicindent} \textbf{Input:} Graph $\mathcal{G}(V,E,w^+,w^-)$; linkage criterion $\interact{}$; \\ 
   \hspace*{4.3em}boolean {\color{blue}addCannotLinkConstraints}  \\
  \hspace*{\algorithmicindent} \textbf{Output:} Final clustering $\Pi^*$, rooted binary hierarchical tree $T^*$\\
  \hspace*{\algorithmicindent} 
  \begin{algorithmic}[1]
  \footnotesize
      \State Initial clustering: $\Pi=\{\{v_1\}, \ldots, \{v_{|V|}\}\}$
      \State Initialize hierarchical tree $T^*$ with leaf nodes  $V=\{v_1,\ldots,v_{|V|}\}$
      \State Initialize  cluster interactions with $\cost_e = w^+_e - w^-_e$, $\forall e\in E$
      \State \emph{// Phase 1: Merge positive interactions (possibly using constraints)}
      \State Push incident nodes of every edge $e\in E$ to priority queue (PQ) with priority $|w_e|$
      \Repeat 
        \State Pop $S,S'\in\Pi$ with highest interaction $|\interact{}(S  \cup  S')|$ from PQ
        \If{\big[{\color{green}\textbf{$\interact{}(S \cup S') > 0$}}\big] \textbf{and} \big[$S,S'$ \textbf{not} constrained\big]}
          \State Merge clusters $S$, $S'$ and update hierarchical tree $T^*$
          \State Update interactions \& constraints with neighboring clusters
        \ElsIf{{\color{blue}addCannotLinkConstr} \textbf{and}  \big[{\color{red}\textbf{$\interact{}(S \cup  S') \leq 0$}}\big]}
          \State Add CannotLink Constraint between $S$ and $S'$
        \EndIf
      \Until{\big[$PQ$ is empty\big]}
      \State \emph{// Phase 2: Remove constraints \& merge all positive interactions}
      \State Push signed interactions $\interact{}(S \cup S')$ to PQ, $\forall S, S' \in \Pi$
      \Repeat 
        \State Pop $S,S'\in\Pi$ with highest interaction $\interact{}(S \cup S')$ from PQ
        \If{\big[{\color{green}\textbf{$\interact{}(S \cup S') > 0$}}\big]}
          \State Merge clusters $S$, $S'$ and update hierarchical tree $T^*$
          \State Update interactions with neighboring clusters
        \EndIf
      \Until{\big[{\color{red}\textbf{$\interact{}(S \cup S') \leq 0$}}\big]} 
      \State Save the final clustering $\Pi^* \gets \Pi$ 
      \State \emph{// Phase 3: Merge negative interactions until one single cluster is left}
      \Repeat
        \State Pop $S,S'\in\Pi$ with highest interaction $\interact{}(S \cup S')$ from PQ 
        \State Merge clusters $S$, $S'$ and update hierarchical tree $T^*$
        \State Update interactions with neighboring clusters
      \Until{\big[Only one cluster is left in $\Pi$\big]} 
      \State
      \Return $\Pi^*$, $T^*$
  \end{algorithmic}
    \label{main_alg}
  \end{flushleft}

\end{algorithm}

\subsection{\algname{}: New and existing algorithms} \label{sec:alg_update_rules}

In this paper we focus on five linkage methods (see columns of Table~\ref{tab:linkage-criteria}). Many more linkage criteria have been applied to unsigned graphs \cite{nunez2013machine,felzenszwalb2004efficient,funke2018large}, involving median-based\footnote{Median linkage is also implemented in our library (see implementation details Appendix 
\ifappendix  
\ref{sec:detailed_impl}).
\else  
A6.1).
\fi
} or size-regularized methods, but we decided to focus this paper on those five criteria because they represent the most popular choices.  

\paragraph{Sum Linkage} -- 
On signed graphs, the sum of two attractive (or repulsive) interactions is still attractive (repulsive). On the other hand, on unsigned graphs, a strong attractive interaction could be obtained by summing many weak interactions, which depending on the application could be undesirable. This explains why, to our knowledge, an agglomerative algorithm with sum linkage has never been used on unsigned graphs. On signed graphs, such an algorithm was pioneered in \cite{levinkov2017comparative,keuper2015efficient} and was named Greedy Agglomerative Edge Contraction (GAEC)\footnote{An algorithm equivalent to GAEC was recently independently re-proposed in \cite{chehreghani2020hierarchical}.}.  
GAEC always makes the \emph{greedy choice} that most decreases the multicut objective defined in Eq.~\ref{eq:MC_objective} each time two clusters with positive interaction are merged\footnote{In general, GASP cannot be seen as a local search algorithm of the multicut problem (for details see Appendix 
\ifappendix  
\ref{sec:relation_to_multicut}).
\else  
A6.2).
\fi
}. The authors of \cite{levinkov2017comparative} propose an algorithm named \emph{GreedyFixation}, which is equivalent to phase one of \algname{} using cannot-link-constraints and a sum linkage. However, running both phase one and two of \algname{} with sum linkage (algorithm named HCC-Sum in this paper) performed better than GreedyFixation in our experiments.

\paragraph{AbsMax Linkage} -- This linkage method is also specific to signed graphs, since on unsigned graphs it would be equivalent to single linkage. Here, we prove that the Mutex Watershed Algorithm \cite{wolf2018mutex} can be seen as an agglomerative algorithm with AbsMax linkage (proofs of the following three propositions are given in Appendix 
\ifappendix  
\ref{sec:proposition_proofs}):
\else  
A6.3): 
\fi
\begin{restatable}{prop}{absmaxmutex}
\label{prop:absmax_mutex}
The \algname{} Algorithm \ref{main_alg} with AbsMax linkage, with or without cannot link constraints, returns the same final clustering $\Pi^*_{\mathrm{AbsMax}}$ also returned by the Mutex Watershed Algorithm (MWS) \cite{wolf2018mutex}, which has empirical complexity $\mathcal{O}(N \log N)$.
\end{restatable}

\paragraph{Average, Single, and Complete Linkage} -- These three linkage criteria have been thoroughly studied on unsigned graphs, but never - until very recently - on signed graphs. In concurrent independent related work \cite{chehreghani2020hierarchical}, the authors prove that applying these three linkage methods to a signed graph is equivalent to applying them to the unsigned graph obtained by shifting all edge weights by a constant.  Here, we prove which of the algorithms studied here are ``intrinsically signed'' and do not have this invariance-property:
\begin{restatable}{prop}{invariantAlgs}
\label{prop:weight_shift_invariant}
We call an agglomerative algorithm ``weight-shift invariant'' if the dendrogram $T$ returned by the algorithm is invariant w.r.t. a shift of all edge weights $w_e$  by a constant $\alpha\in \mathbb{R}$. Among the variations of \algname{}, only hierarchical clustering with Average (HC-Avg), Single (HC-Single), and Complete linkage (HC-Complete) are weight-shift-invariant (see green box in Table~\ref{tab:linkage-criteria}).
\end{restatable}
\noindent Although average and single linkage methods have been largely studied on unsigned graphs, to our knowledge, they have never been combined with cannot-link constraints on signed graphs\footnote{Note that Complete linkage methods return the same clustering whether constraints are enforced or not (proof in Lemma 
\ifappendix 
\ref{lemma:absMax_and_complete_property}, 
\else  
A6.3, 
\fi 
in Appendix).}, so we name these algorithms \emph{HCC-Avg} and \emph{HCC-Single}.

\paragraph{Algorithms defining an ultrametric} -- 
The connection between agglomerative algorithms and ultrametrics\footnote{A metric space $(X,d)$ is an \emph{ultrametric} if, for every $x,y,z \in X$, $d(x,y)\leq \max \{d(x,z), d(y,z)\}$.} is well known. Usually, ultrametrics are associated to strictly positive \emph{similarity} or \emph{dissimilarity} measures on a graph. In our framework, a trivial ultrametric is always given by the height $h_T$ of the dendrogram. However, for some of the \algname{} variations, we now define an ultrametric based on the edge weights and the signed interactions between clusters, generalizing what has been done for HC on unsigned graphs \cite{johnson1967hierarchical,milligan1979ultrametric}. To define this measure and prove its ultrametric property, we first map the signed interaction $\treeHeight{}$ defined in Eq.~\ref{eq:def_dendr_interact} to positive ``pseudo-distances'' $d_{T}:V \times V \rightarrow \mathbb{R}^{+}$:
\begin{align}\label{eq:UM_def}
d_{T}(u,v)\equiv &\begin{cases}
0 & \text{if}\,\, u=v\\
M-\treeHeight(u,v) & \text{if}\,\, u\neq v
\end{cases}\quad \forall u,v\in V\\
\mathrm{where}& \quad  M \equiv  \epsilon + \max_{u',v'\in V,\,u'\neq v'}\treeHeight(u',v')
\end{align}
and where $\epsilon > 0$. 
We then prove the following proposition:
\begin{restatable}{prop}{secondUltraMetricProperty}
\label{prop:ultraMetric2}
Among the algorithms included in the \algname{} framework (see Table \ref{tab:linkage-criteria}), only Mutex Watershed and hierarchical clustering with Average (HC-Avg), Single (HC-Single) and Complete linkage (HC-Complete) define an ultrametric $(V, d_{T^*})$, where $d_{T^*}$ is defined in Eq.~\ref{eq:UM_def} and $T^*$ is the tree returned by the \algname{} Algorithm~\ref{main_alg}.
\end{restatable}
\noindent In summary, in this section we have extended the family of HC algorithms \cite{johnson1967hierarchical,milligan1979ultrametric} with ``weight-based ultrametrics'' to signed graphs. Next, we move to their empirical evaluation.

%% file: chapters/5_experiments_intro.tex
\begin{table*}[t]
    \centering
    \scriptsize
    \begin{subtable}[t!]{\textwidth}
    \centering

        \begin{tabular}{l  c  r  c  c  | r r r r r r}
        &&&& %
        &\multicolumn{5}{c}{Multicut objective values (average across instances, lower is better)} \\
        Clustering problem & \makecell{Graph Type} & $\#I$ & $|V|$ & $|E|$  & \multicolumn{1}{r}{GAEC \cite{keuper2015efficient}} & HCC-Sum & MWS \cite{wolf2018mutex} & HC-Avg & HCC-Avg \\ \midrule
        \emph{Modularity Clustering} \cite{brandes2007modularity} & \emph{complete} & 6& 34-115 & 561-6555 & %
        -0.457 & -0.453 & -0.073 & \textbf{-0.467} & \textbf{-0.467} \\ 
        \emph{Image Segmentation} \cite{andres2011probabilistic} & \emph{RAG} & 100 & 156-3764 &  439-10970  & %
        \textbf{-2,955} & -2,953 & -2,901 & -2,903 & -2,896\\
        \emph{Knott-3D (150-300-450)} \cite{andres2012globally} & \emph{3D-RAG} & 24 & 572-17k & 3381-107k & %
        \textbf{-36,667} & -36,652 & -35,200 & -35,957 & -35,631\\
        \emph{CREMI-3D-RAG (OurCNN)}  & \emph{3D-RAG} & 3& 134k-157k & 928k-1065k %
        & \textbf{-1,112,287} & \textbf{-1,112,286}& -1,109,731 & -1,112,177 & -1,112,100\\ 
        \emph{Fruit-Fly Level 1-4} \cite{pape2017solving} & \emph{3D-RAG} & 4& 5m-11m & 28m-72m %
        & \textbf{-151,022} & -151,017 & -150,879 & -150,909 & -150,876\\
        \emph{CREMI-gridGraph (OurCNN)} & \emph{gridGraph} & 15& 39m & 140m %
        & -73,317,601 & -73,328,867 & -73,330,568 & \textbf{-73,502,947} & -73,474,856\\
        \emph{Fruit-Fly Level Global} \cite{pape2017solving} & \emph{3D-RAG} & 1& 90m & 650m %
        & \textbf{-151,688} & -151,596 & -146,315 & -150,466 & -150,171 \\

        \end{tabular}
    \end{subtable} 
    \caption{List of compared signed graph clustering problems: for each, we specify the number of instances $\# I$, number of nodes $|V|$, and number of edges $|E|$ per instance. We compare algorithms in the \algname{} framework by their value of the multicut objective defined in Eq.~\ref{eq:MC_objective} (lower is better).} 
    \label{tab:datasets_and_energies}
\end{table*}

\section{Experiments}\label{sec:neuro_segm_exp}
\subsection{Signed graph clustering problems} \label{sec:clustering_problems}
We evaluate the agglomerative clustering algorithms included in our framework on a large collection of both synthetic and real-world graphs with very different structures. The size of the graphs ranges from a few hundred to hundreds of millions of edges. In this way, we will highlight the strengths and limitations of the different linkage criteria introduced in the last section. 

\textbf{Synthetic SSBM graphs} -- We first consider synthetic graphs generated by a signed stochastic block model (SSBM). We use an Erd\H os-R\'enyi random graph model $\mathcal{G}(N,p)$ with $N$ vertices and edge probability $p$. Following the approach in \cite{Cucuringu2019SPONGEAG}, we partitioned the graph into $k$ \emph{ground-truth} clusters, such that edges connecting vertices belonging to the same cluster (different clusters, respectively) have Gaussian distributed edge weights centered at $\mu=1$ ($\mu=-1$, respectively) and with standard deviation $\sigma=0.1$. To model noise, the sign of the edge weights is flipped independently with probability $\eta$.

\textbf{Existing signed graphs}  -- We use clustering instances from the OpenGM benchmark \cite{kappes2013comparative} as well as biomedical segmentation instances \cite{pape2017solving}. The dataset \emph{Image Segmentation} contains planar region-adjacency-graphs (RAG) that are constructed from superpixel adjacencies of photographs. The \emph{Knott-3D} datasets contains 3D-RAGs arising from volume images acquired by electron microscopy (EM). The set \emph{Modularity Clustering} contains complete graphs constructed from clustering problems on small social networks. The \emph{Fruit-Fly} 3D-RAG instances were generated from volume image scans of fruit fly brain matter. Instances \emph{Level 1-4} are progressively simplified versions of the global problem obtained via block-wise domain decomposition \cite{pape2017solving}.

\textbf{Grid-graphs from CNN predictions} -- We also evaluate the clustering methods on the task of neuron segmentation in EM image volumes using training data from the CREMI 2016 EM Segmentation Challenge \cite{cremiChallenge}.
We train a 3D U-Net \cite{ronneberger2015u,cciccek20163d} using the same architecture as \cite{funke2018large} and predict long-and-short range affinities 
as described in \cite{lee2017superhuman}. The predicted affinities $a_e\in[0,1]$, which represent how likely it is for a pair of pixels to belong to the same neuron segment, are then mapped to signed edge weights $w_e=a_e-0.5$, resulting in a 3D grid-graph having a node for each pixel/voxel of the  image\footnote{To map affinities to signed weights, we also tested the \emph{logarithmic mapping} proposed in \cite{finkel2008enforcing,andres2012globally}, but it performed worse in our experiments.}. 
We divided the three CREMI training samples, consisting of $\sim$196 million voxels each, into five sub-blocks for a total of 15 clustering problems (named \emph{CREMI-gridGraph} in Table~\ref{tab:datasets_and_energies}). See Appendix 
\ifappendix  
\ref{sec:cremi_details} 
\else  
A6.5 
\fi
 for extended details about training, data augmentation, and how we remove tiny clusters left after running \algname{} on the \emph{CREMI-gridGraph} clustering problems.

\textbf{3D-RAG from CNN-predictions} -- Lastly, we use the predictions of our CNN model to generate three graph instances (one for each CREMI training sample, named \emph{CREMI-3D-RAG} in Table~\ref{tab:datasets_and_energies}), which have very similar structure to the \emph{Knott-3D} and \emph{Fruit-Fly} instances.  We obtain these problems by using a pipeline that is very common in neuron segmentation: a watershed algorithm generates superpixels and from those a 3D region-adjacency graph is built, where edge weights are given by the CNN predictions averaged over the boundaries of adjacent superpixels (details in Appendix 
\ifappendix  
\ref{sec:cremi_details}). 
\else  
A6.5). 
\fi

\begin{table*}[t]
        \centering
    \tiny
        \begin{subtable}[t]{0.27\textwidth}
        \centering
        \begin{tabular}[t]{@{\hspace{0.7\tabcolsep}}l c @{\hspace{1\tabcolsep}} c @{\hspace{1.1\tabcolsep}} c @{\hspace{1\tabcolsep}} c @{\hspace{1\tabcolsep}}}
        \toprule
          & ARAND & VOI & VOI&  Runtime \\ 
          & Error & split & merge&  (s) \\ \midrule 
\textbf{HC-Avg} & \textbf{0.0487} & 0.387 & 0.258 & 2344 \\
HCC-Avg & 0.0492 & 0.389 & 0.259 & 2892 \\
MWS \cite{wolf2018mutex} & 0.0554 & 0.440 & 0.249 & 688 \\
GAEC \cite{keuper2015efficient} & 0.0856 & 0.356 & 0.338 & 4717 \\
HCC-Sum & 0.0872 & 0.365 & 0.337 & 4970 \\
HC-Complete & 0.9211 & 4.536 & \textbf{0.211} & 1020 \\
HC-Single & 0.9264 & \textbf{0.060} & 4.887 & \textbf{312} \\
HCC-Single & 0.9264 & \textbf{0.060} & 4.887 & 6440 \\
        \end{tabular}
        \vspace*{1.1em}
    \caption{\emph{CREMI-gridGraph (OurCNN)}}
    \label{tab:scores_gridGraph}
    \centering
    \tiny
    \end{subtable}\hfill
\begin{subtable}[t]{.27\textwidth}
\centering
        \begin{tabular}[t]{@{\hspace{0.7\tabcolsep}}l c @{\hspace{1\tabcolsep}} c @{\hspace{1.1\tabcolsep}} c @{\hspace{1\tabcolsep}} c @{\hspace{1\tabcolsep}}}
        \toprule
          & ARAND & VOI & VOI&  Runtime \\ 
          & Error & split & merge&  (s) \\ \midrule 
\textbf{HC-Avg} & \textbf{0.0896} & 0.603 & 0.323 & 86 \\
HCC-Avg & 0.0898 & 0.600 & 0.325 & 87 \\
GAEC \cite{keuper2015efficient} & 0.0905 & 0.606 & 0.323 & 89 \\
HCC-Sum & 0.0910 & 0.608 & 0.323 & \textbf{85} \\
MWS \cite{wolf2018mutex} & 0.1145 & 0.825 & 0.295 & 86 \\
HCC-Single & 0.5282 & \textbf{0.437} & 1.367 & 88 \\
HC-Single & 0.5282 & \textbf{0.437} & 1.367 & \textbf{85} \\
HC-Complete & 0.5654 & 2.253 & \textbf{0.249} & 86 \\
        \end{tabular}
        \vspace*{1.1em}
    \caption{\emph{CREMI-3D-RAG (OurCNN)}}
    \label{tab:scores_3drag}
    \end{subtable} \hfill
    \begin{subtable}[t]{.43\textwidth}
    \centering
    \tiny
        \begin{tabular}[t]{l @{\hspace{1.2\tabcolsep}} c @{\hspace{1\tabcolsep}} c @{\hspace{1\tabcolsep}} c @{\hspace{0.8\tabcolsep}} c @{\hspace{1\tabcolsep}} c}
        \toprule
        & Needs & CREMI& ARAND & VOI & VOI\\ 
          & superpixels? & Score & Error & split & merge\\ \midrule 
OurCNN: 3D-RAG + LiftedMulticut & \CrossedBox & \textbf{0.221} & \textbf{0.108} & \textbf{0.339} & 0.115 \\
\emph{GASP: OurCNN + gridGraph + HCC-Avg} & \HollowBox & 0.224 & 0.113 & 0.361 & 0.085  \\
\emph{GASP: OurCNN + gridGraph + HC-Avg}  & \HollowBox &0.224 & 0.114 &  0.364 & 0.083 \\
PNI CNN \cite{lee2017superhuman} & \CrossedBox &0.228 & 0.116 & 0.345 & 0.106 \\
LSI-Masks \cite{bailoni2020proposal}  & \HollowBox &0.246 & 0.125 & 0.383 & 0.107  \\
\emph{GASP: OurCNN + 3D-RAG + HCC-Avg} & \CrossedBox &0.257 & 0.132 & 0.438& \textbf{0.063} \\  
\emph{GASP: OurCNN + 3D-RAG + HC-Avg} & \CrossedBox &0.262 & 0.135 & 0.448 & \textbf{0.063}   \\  
MALA CNN + MC \cite{funke2018large} & \CrossedBox & 0.276  & 0.132 &0.490  & 0.089  \\
CRU-Net \cite{zeng2017deepem3d} & \CrossedBox &0.566 & 0.229 & 1.081 &  0.389    \\
        \end{tabular}
        \caption{CREMI Challenge leader-board}
        \label{tab:cremi_leaderboard}
        \end{subtable}
    \caption{\textbf{Tables (a-b)}: Scores and run times of algorithms in the \algname{} framework on the \emph{CREMI-gridGraph} and \emph{CREMI-3D-RAG} clustering problems: average linkage methods achieved the best accuracy. Measures shown are: Adapted-Rand error (ARAND, lower is better); Variation of Information (VOI) \cite{arganda2015crowdsourcing} (VOI-merge for under-clustering error and VOI-split for over-clustering error, lower values are better). \textbf{Table (c)}: Current leading entries in the CREMI challenge leaderboard (November 2021). CREMI-score is given by the geometric mean of (VOI-split + VOI-merge)  and ARAND error (lower is better).}
    \label{tab:scores}
\end{table*}

\subsection{Comparison of results and discussion}\label{sec:experiments_discussion}

\textbf{Multicut objective values} -- In Table~\ref{tab:datasets_and_energies}, we report the values of the multicut objective obtained for clustering with different \algname{} algorithms\footnote{Objective values achieved by Single and Complete linkage methods are much worse compared to other algorithms and are reported in Table 
\ifappendix  
\ref{tab:all_multicut_energies}, 
\else  
A5, 
\fi
 in Appendix.}. Although many heuristics were proposed to better optimize this objective \cite{beier2016efficient,beier2014cut,kernighan1970efficient}, these methods are out of the scope of this paper, since they do not scale to the largest graph instances considered here. By looking at results in Table~\ref{tab:datasets_and_energies}, we observe that GAEC almost always achieves the lowest objective values, expect in the \emph{CREMI-gridGraph} instances. Despite this, on graphs where a ground truth clustering is known, GAEC does not achieve the lowest ARAND errors (see Tables~\ref{tab:scores_gridGraph} and \ref{tab:scores_3drag}). 

\begin{figure}
\centering
\includegraphics[width=0.48\textwidth,trim=0 10 60 0, clip]{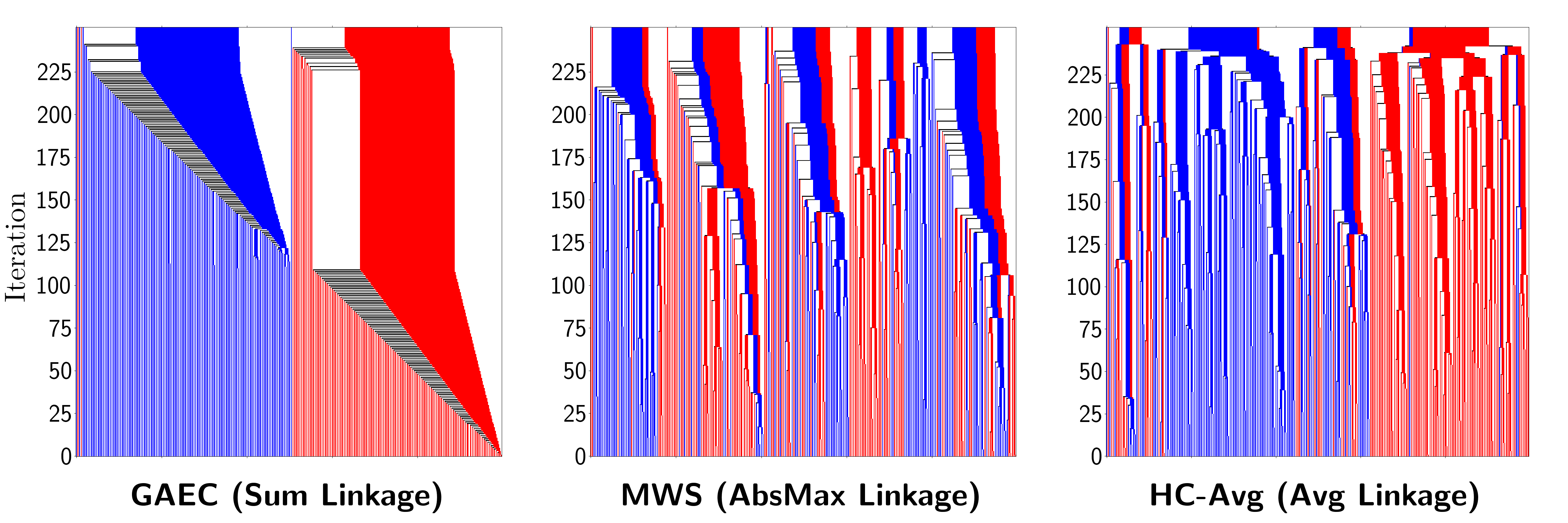} %
\caption{Clustering dynamics and accuracy of GASP variations on stochastic block models. The dendrograms result from three versions of \algname{} on a synthetic graph generated with SSBM ($250$ nodes, edge probability $p=0.05$, flipping probability $\eta=0.1$).  Red and blue colors show which of the two equal-sized ground-truth communities each node belongs to. At the top, dendrograms are truncated at the level of the final clustering $\Pi^*$ returned by \algname{}. \label{fig:dendrograms}}
\end{figure}

\textbf{Size of growing clusters: Sum vs Avg linkage} --
In all the studied clustering problems, we empirically observe that sum-linkage algorithms like GAEC grow clusters one after the other, as shown in Fig.~\hyperref[fig:intro_figure]{\ref*{fig:intro_figure}d} and Fig.~\ref{fig:dendrograms} by the agglomeration order of GAEC\footnote{This \emph{flooding agglomeration-strategy} of GAEC was also observed in \cite{kardoostsolving}.}. This is intuitively explained by the  following: initially, many of the most attractive edge weights have very similar values; when the two nodes $u,v$ with the highest attraction are merged, there is a high chance that they will have a common neighboring node $t$ belonging to the same cluster; thus, the interaction between the merged nodes $uv$ and $t$ is likely assigned to the highest priority, because it is given by the sum of two highly attractive edge weights. This will then start a ``chain reaction'' where only a single cluster is agglomerated at the time. 
In the following, we will show how this unique \emph{flooding strategy} of the sum-linkage methods can be both an advantage or a disadvantage, depending on the type of clustering problem. 

\begin{figure}[t]
\begin{subfigure}[t]{0.23\textwidth}
\centering
\includegraphics[width=\textwidth,trim=0.34in 0.34in 0.34in 0.34in,clip]{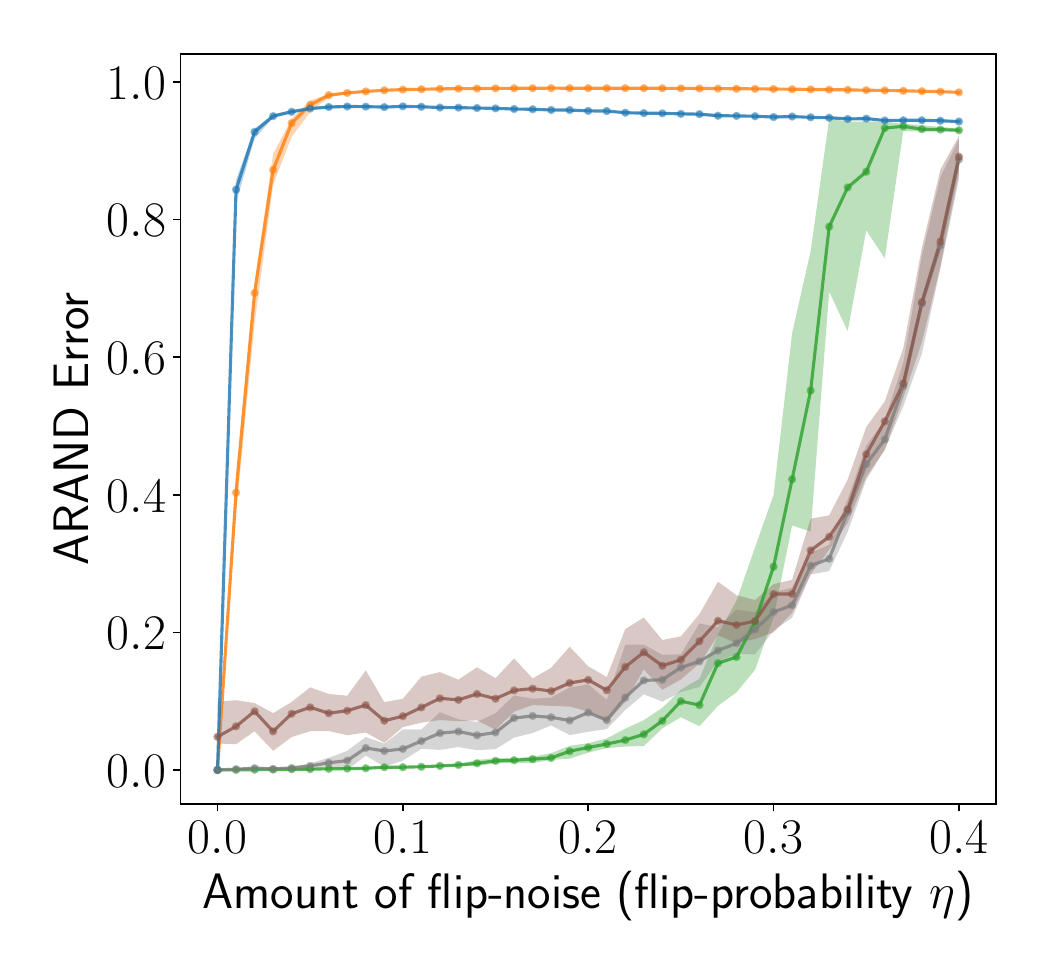}
\caption{$k=20$, $p=0.1$}
\end{subfigure}\hfill
\begin{subfigure}[t]{0.23\textwidth}
\centering
\includegraphics[width=\textwidth,trim=0.34in 0.34in 0.34in 0.34in,clip]{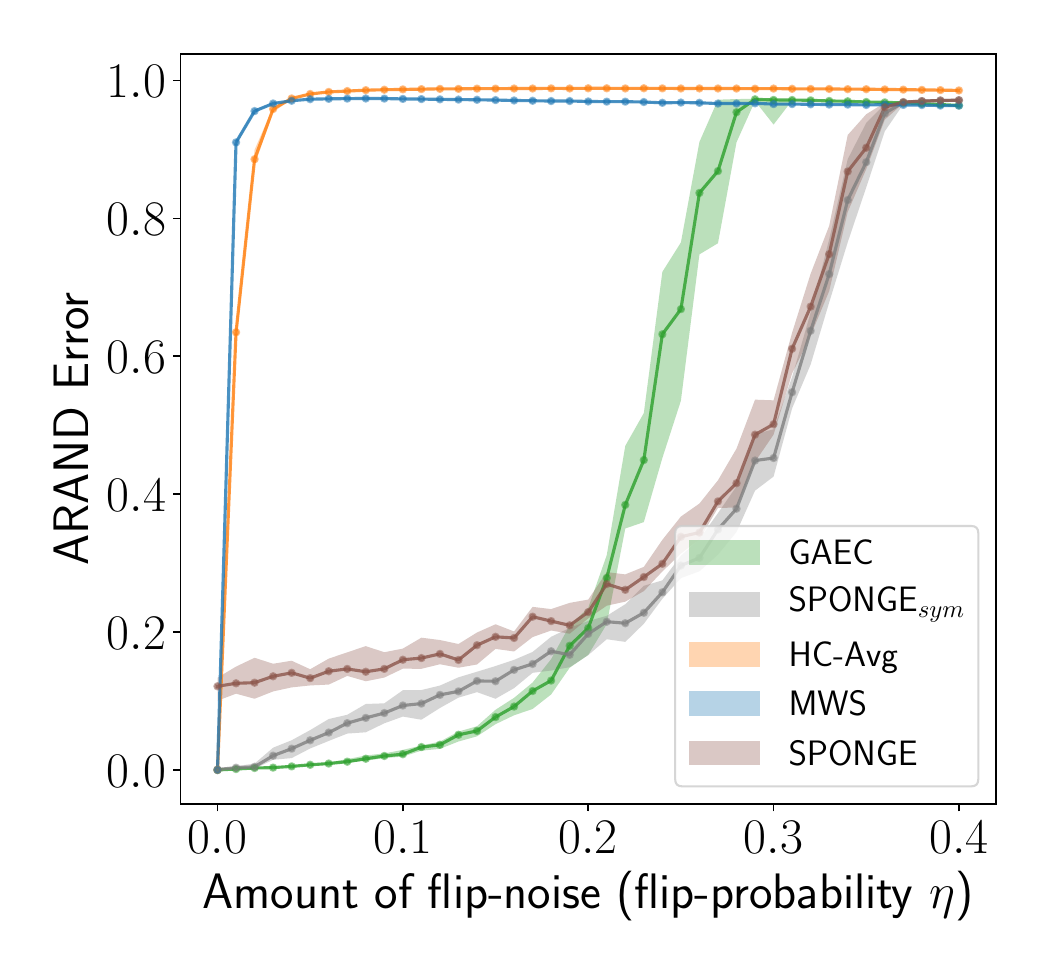}
\caption{$k=50$, $p=0.2$}
\end{subfigure}
        \caption{
ARAND errors (median values over 20 experiments, lower is better) on synthetic graphs generated with SSBM. We consider $k$ ground truth communities of random size. Graphs have $N=10000$ nodes; edges are randomly added with probability $p$. 
        } \label{fig:SSBM_scores}
\end{figure}

\textbf{Comparison to spectral clustering} -- 
The spectral clustering methods for signed graphs SPONGE$_{sym}$ and SPONGE proposed by \cite{Cucuringu2019SPONGEAG} achieved state of the art performances on SSBM synthetic graphs. Their competitive performances are also confirmed by our experiments in Fig.~\ref{fig:SSBM_scores}.
However, these methods do not scale up to the large graph instances considered here and they also require the user to specify the true number of clusters in advance, which is not known for other graph instances tested in this paper. In Appendix, Table 
\ifappendix  
\ref{tab:cremi_spectral_experiments}, 
\else  
A6, 
\fi
 we report the scores achieved by these methods on a much smaller sub-instance of the \emph{CREMI-gridGraph} problem: even when the true number of clusters is specified in advance for the spectral methods, they perform much worse than other \algname{} algorithms, with an accuracy penalty of almost 50\%. For these reasons, we exclude them from our other comparison experiments.  

\textbf{GASP on synthetic SSBM graphs}  -- 
\algname{} algorithms using cannotLinkConstraints are not expected to perform well on these graphs, because of the type of employed sign noise, so we focus our comparison only on the GAEC, HC-Avg and MWS algorithms (using Sum, Average, and AbsMax linkage methods, respectively).
Empirically, we observe that GAEC is the agglomerative algorithm performing best on SSBM graphs, on par with spectral method SPONGE$_{sym}$ (see Fig.~\ref{fig:SSBM_scores}). Given the simple properties of SSBM graphs, we can now give a detailed explanation of these empirical results. 
In SSBM graphs, the number of edges $E_{ij}$
connecting two clusters $S_i,S_j$ is  proportional to the product $|S_i|\cdot|S_j|$ of cluster sizes. 
With Sum or Avg linkage methods, due to the law of large numbers, the flipping noise is ``averaged out'' as soon as the set $E_{ij}$ becomes larger and clusters grow in size.
On the other hand, when clusters are small, it can happen that, for few clusters, several of their edges in $E_{ij}$ are flipped and the algorithm makes a mistake by merging two clusters belonging to different ground truth communities. From this observation, it follows that the \emph{flooding strategy} of the sum-linkage algorithm GAEC is a very good strategy on these types of graphs, because clusters are immediately grown in size (see dendrograms in Fig.~\ref{fig:dendrograms}). Average linkage method HC-Avg instead performs much worse on these graphs because it grows small equally-sized clusters and makes several wrong merge-decisions at the beginning. 
Lastly, the MWS algorithm is not expected to perform well on these graphs because of the high sensitivity of the AbsMax linkage to flipping noise. In Proposition 
\ifappendix  
\ref{prop:MWS_on_SSBM} 
\else  
A6.2 
\fi
 (see Appendix), we prove that, at every iteration, the MWS algorithm makes a mistake with at least probability $\eta$, independently on the sizes of the two clusters that are popped from priority queue.
In summary, for the SSBM, we can obtain a deep understanding of the dynamics induced by various linkage criteria, and find that GAEC gives highest accuracy by a large margin.

\begin{figure}
\centering
\includegraphics[width=0.48\textwidth,trim=0 10 60 0, clip]{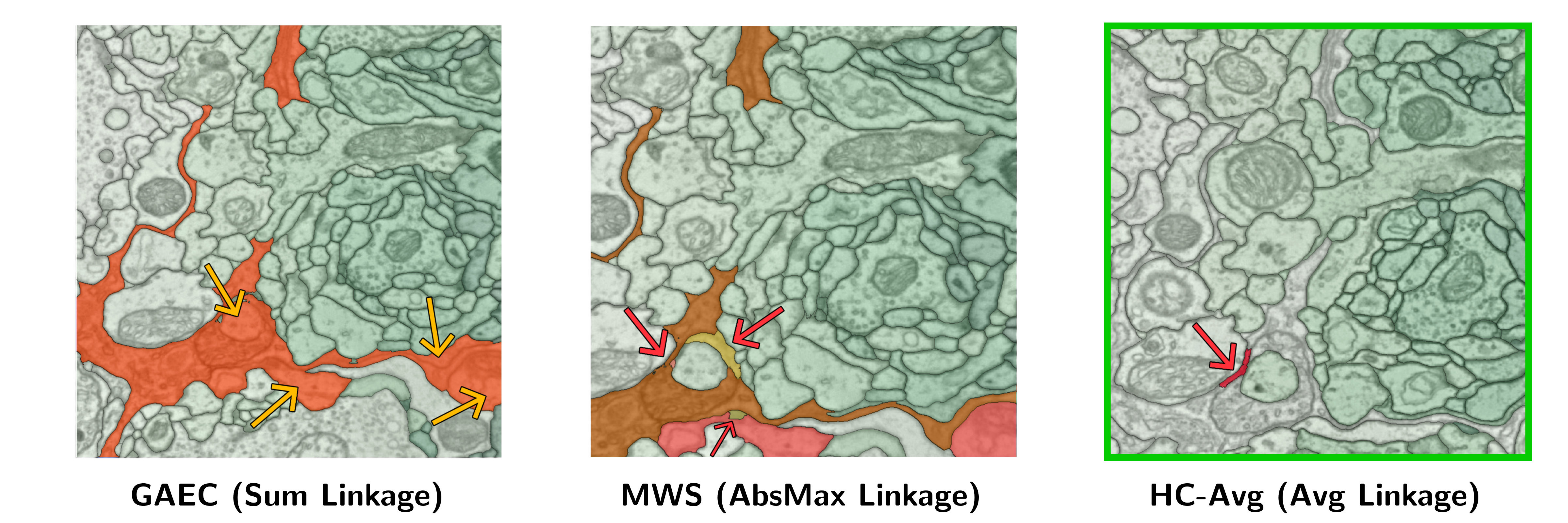} %
\caption{Failure cases of three versions of \algname{} applied to neuron segmentation. Only \emph{wrongly} segmented regions are highlighted in different warm colors. Red arrows point to wrongly split regions; yellow arrows point to false merge errors. HC-Avg returned the best segmentation. Data is 3D, hence the same color could be assigned to parts of segments that appear disconnected in 2D.  
\label{fig:failure_cases}}
\end{figure}

\textbf{Scores on CREMI instance-segmentation} -- 
SSBM graphs are \emph{non-planar}, and every edge has the same probability to be present in the graph. On the other hand, the \emph{gridGraph} and \emph{3D-RAG} graphs of Table~\ref{tab:datasets_and_energies} are sparse and have a very regular structure: regardless of whether a node represents a pixel or a superpixel, it will only have edge connections with its neighbors in the image (up to a certain hop distance). 
Tables~\ref{tab:scores_gridGraph}-\ref{tab:scores_3drag} show that average linkage methods (HC-Avg, HCC-Avg) strongly outperform other methods on \emph{CREMI-gridGraph} instances and also achieve the best scores on \emph{CREMI-3D-rag} graphs. Sum-based linkage methods (GAEC, HCC-Sum) have a two times higher ARAND error on grid-graphs and often return under-clustered segments (see failure cases in Fig.~\ref{fig:failure_cases}). This suggests that the \emph{flooding strategy} observed previously in the sum-linkage methods does not work on grid-graphs, because in this setup edge weights are predicted by a CNN and noise is strongly spatially-correlated \footnote{This effect is not as strong on \emph{3D-RAG} graphs, because edge weights are computed by averaging CNN predictions (and noise) over the boundaries of adjacent supervoxels.}.
To fully test this hypothesis, we conduct a set of experiments where the CNN predictions are perturbed by adding structured noise and simulating additional artifacts like ``holes'' in the boundary evidence\footnote{See Appendix 
\ifappendix  
\ref{sec:appendix_noise_gen} 
\else  
A6.6 
\fi
 for details about how we perturbed the \emph{CREMI-gridGraph} problems by using Perlin noise \cite{perlin2001noise,perlin1985image}, which is one of the most common gradient noises used in procedural pattern generation.}. 
The plot in Fig.~\ref{fig:scores_structured_noise} confirms that HC-Avg and HCC-Avg are very robust algorihtms on this data, followed by Sum-linkage algorithms and the Mutex Watershed algorithm (MWS). It is not a surprise that the AbsMax linkage used by MWS is not robust to this type of structured noise. However, the scores and runtimes in Table~\ref{tab:scores_gridGraph} prove how MWS can achieve high accuracy with 70\% lower runtime compared to HC-Avg. 

\textbf{Complete and Single Linkage} -- We use these two linkage methods as baselines to highlight the difficulty of the studied graph clustering problems listed in Table~\ref{tab:datasets_and_energies}. Scores in Tables~\ref{tab:scores_gridGraph}-\ref{tab:scores_3drag} show their poor performance: Single linkage hierarchical clustering (HC-Single), which here is equivalent to thresholding the edge weights at $w_e=0$ and computing connected components in the graph, often returned few big under-segmented clusters. HC-Complete returned instead a lot of over-segmented clusters. 

\textbf{Results on CREMI challenge} -- 
Table~\ref{tab:cremi_leaderboard} shows that the HCC-Avg and HC-Avg clustering algorithms achieve state-of-the-art accuracy on the CREMI challenge, when combined with predictions of our CNN.
Most of the other entries (apart from \emph{LSI-Masks} \cite{bailoni2020proposal}) employ super-pixels based post-processing pipelines and cluster 3D-region-adjacency graphs. As we show in Table ~\ref{tab:scores_3drag}, using superpixels considerably reduces the size of the clustering problem and, consequently, the post-processing time. 
However, our method operating directly on pixels (\emph{gridGraph + HCC-Avg}) achieves better performances than superpixel-based methods (\emph{3D-RAG + HCC-Avg}) and does not require the parameter tuning necessary to obtain good super-pixels, which is usually highly dataset dependent.
To scale up our method operating on pixels, we divided each test-volume into four sub-blocks, and then combined the resulting clusterings by running the algorithms again on the combined graph.
The method \emph{3D-RAG + LiftedMulticut} based on the lifted multicut approximation of \cite{beier2017multicut} achieves the best scores overall, but it takes into account different information through the lifted edge weights that also depend on additional raw-data and shape information from highly engineered super-pixels.

%% file: chapters/5_experiments_compare_rules.tex
\begin{figure}[b]
\centering
\includegraphics[width=0.47\textwidth,trim=0.2in 0.17in 0.2in 0.2in,clip]{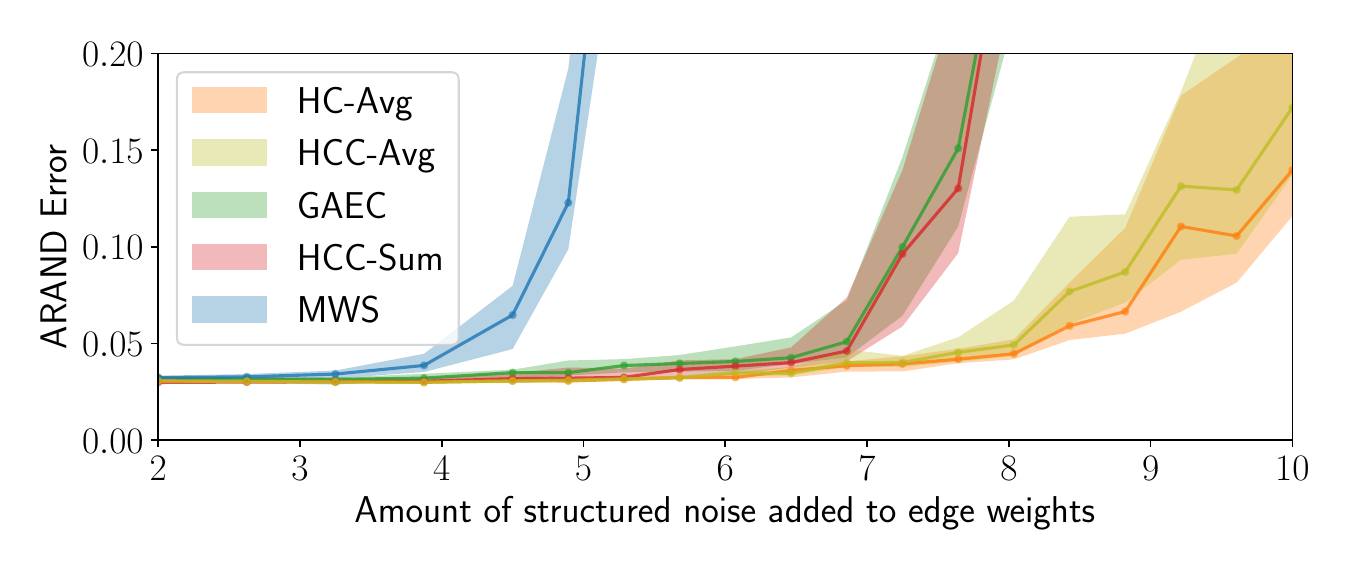}
        \caption{
ARAND errors (median values over 20 experiments, lower is better) on \emph{CREMI-gridGraph} clustering problems perturbed with structured noise. Average-linkage algorithms proved to be the most robust.
}\label{fig:scores_structured_noise}
\end{figure}

%% file: chapters/6_conclusions.tex
\section{Conclusion}
We have presented a unifying framework for agglomerative clustering of graphs with both positive and negative edge weights. This framework allowed us to explore new combinations of constraints and linkage criteria and to perform a consistent evaluation of all algorithms in it. 
We have then analyzed several theoretical and empirical properties of these algorithms. On instance segmentation, algorithms based on an average linkage criterion outperformed all the others: they proved to be simple and robust approaches to process short- and long-range predictions of a CNN.
On biological images, these simple average agglomeration algorithms achieve state-of-the-art results without requiring the user to spend much time tuning complex task-dependent pipelines based on super-pixels.

%% file: chapters/supplementary.tex
\renewcommand{\thesection}{A\arabic{section}}
\renewcommand{\thetable}{A\arabic{table}}
\renewcommand{\thefigure}{A\arabic{figure}}

\section{Appendix}

\subsection{Implementation and complexity of \algname{}} \label{sec:detailed_impl}

\begin{figure}[t]
        \centering
        \includegraphics[width=0.48\textwidth]{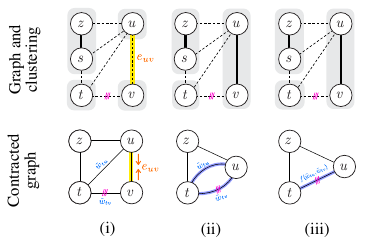} %
    \centering
    \caption{Example of edge contraction. First row: original graph $\mathcal{G}$; clustering $\Pi$ (gray shaded areas) with dashed edges on cut; cannot-link constraints (violet bars). Second row: contracted graph $\tilde{\mathcal{G}}_\Pi$. In step ii), edge $e_{uv}$ is contracted and node $v$ deleted from $\tilde{\mathcal{G}}_\Pi$. In step iii), double edges $e_{tu}$ and $e_{tv}$ resulting from the edge contraction are replaced by a single edge with updated interaction.}\label{fig:edge_contraction_and_contr_graph}  
\end{figure}

\paragraph*{Update rules} During the agglomerative process, the interaction between adjacent clusters has to be properly updated and recomputed, as shown in Algorithm \ref{main_alg}.  
An efficient way of implementing these updates can be achieved by representing the agglomeration as a sequence of \emph{edge contractions} in the graph. Given a graph $\mathcal{G}(V,E,\cost)$ and a clustering $\Pi$, we define the associated \emph{contracted graph} $\tilde{\mathcal{G}}_\Pi(\tilde{V}, \tilde{E}, \tilde{\cost})$, such that there exists exactly one representative node $|\tilde{V} \cap S| = 1$ for every cluster $S \in \Pi$ . Edges in $\tilde{E}$ represent adjacency-relationships between clusters 
and the signed edge weights $\tilde{\cost}_e$ are given by inter-cluster interactions $\tilde{\cost}(e_{uv})=\interact_{S_u \cup S_v}$, where $S_u$ denotes the clustering including node $u$. 
For the linkage criteria tested in this article, when two clusters $S_u$ and $S_v$ are merged, the interactions between the new cluster $S_u \cup S_v$ and each of its neighbors depend only on the previous interactions involving $S_u$ and $S_v$. Thus, we can recompute these interactions by using an \emph{update rule} $f$ that does not involve any loop over the edges of the original graph $\mathcal{G}$:
\begin{align}
  \interact(S_u \cup S_v  \cup S_t) =& f\Big[ \interact(S_u  \cup  S_t), \interact(S_v  \cup  S_t) \Big] \\
  =& f(\tilde{\cost}(e_{ut}), \tilde{\cost}(e_{vt})) 
\end{align}
In Fig. \ref{fig:edge_contraction_and_contr_graph} we show an example of edge contraction and in Table \ref{tab:linkage_criteria_explicit} we list the update rules associated to the linkage criteria we introduced in Table \ref{tab:linkage-criteria}.

\afterpage{ %
  \begin{algorithm*}[p]
    \caption{Implementation of \algname{} - Phase 1}
  \hspace*{\algorithmicindent} \textbf{Input:} $\mathcal{G}(V,E,w^+,w^-)$ with $N$ nodes and $M$ edges; boolean \texttt{{\color{blue}addCannotLinkConstraints}} \\
  \hspace*{\algorithmicindent} \textbf{Output:} Final clustering \\
    \hspace*{\algorithmicindent} 
    \begin{algorithmic}[1]
        \State $\tilde{\mathcal{G}}(\tilde{V},\tilde{E}) \gets \mathcal{G}(V,E,w^+,w^-)$  \Comment{Init. contracted graph}
        \State \texttt{UF} $\gets$ initUnionFind($V$) \Comment{Init. data structure representing clustering}
          \State PQ.push$(|w_e|, e) \quad \forall e \in E $  \Comment{Init. priority queue in desc. order of $|w_e|=|w_e^+ - w_e^-|$, $\mathcal{O}(|E|)$}
          \State \texttt{canBeMerged}$[e] \gets$ \texttt{True} $\,\,\, \forall e\in E$ \Comment{Init. cannot-link constraints}
      \State
        \While{PQ is \textbf{not} empty}
          \State $\tilde{w}, e_{uv} \gets $ PQ.popHighest() \Comment{$\mathcal{O}(\log |E|)$}
          \State \textbf{assert} \texttt{UF}.find($u$) $\neq$ \texttt{UF}.find($v$) \Comment{Edges in PQ always link nodes in different clusters}
          \If{({\color{green}\textbf{$\tilde{w} > 0$}}) \textbf{and} \texttt{canBeMerged}$[e_{uv}]$}
            \State PQ, \texttt{canBeMerged}, $\tilde{E}$ $\gets$ \textsc{UpdateNeighbors}($u,v$)
            \State $\tilde{V} \gets \tilde{V} \setminus \{ v\}$, $\quad \tilde{E} \gets \tilde{E} \setminus \{ e_{uv}\}$ \Comment{Update contracted graph}
            \State \texttt{UF}.merge($u,v$) \Comment{Merge clusters, $\mathcal{O}(\alpha(|E|))$}
          \ElsIf{({\color{red}\textbf{$\tilde{\cost} \leq 0$}}) \textbf{and} {\color{blue}\texttt{addCannotLinkConstraints}}}
            \State \texttt{canBeMerged}$[e_{uv}] \gets$ \texttt{False} \Comment{Constrain the two clusters}
          \EndIf
        \EndWhile
        \State
        \Return Final clustering given by union-find data structure  \texttt{UF}
    \end{algorithmic}
    \hspace*{1.5cm} 
      \begin{algorithmic}[1]
      \Function{UpdateNeighbors}{$u,v$}
        \State $\mathcal{N}_u = \{ t \in \tilde{V} | e_{ut}\in \tilde{E}  \}$
        \State $\mathcal{N}_v = \{ t \in \tilde{V} | e_{vt}\in \tilde{E}  \}$ 
        \For{$t \in \mathcal{N}_v$ } \Comment{Loop over neighbors in $\tilde{\mathcal{G}}$ of deleted node $v$}
          \State $\tilde{E} \gets \tilde{E} \setminus \{e_{vt}\}$
          \State $\tilde{w}_{vt} \gets$ PQ.delete($e_{vt}$) \Comment{Delete edge $e_{vt}$ from PQ and get the old edge weight, $\mathcal{O}(\log |E|)$}
          \State \texttt{canBeMerged}$[e_{ut}] \gets$ \texttt{canBeMerged}$[e_{ut}]$ \textbf{and} \texttt{canBeMerged}$[e_{vt}]$
          \If{$t \in \mathcal{N}_u$ }\Comment{Check if $t$ is a common neighbor of $u$ and $v$}
            \State $\tilde{w}_{ut} \gets$ PQ.delete($e_{ut}$)  \Comment{$\mathcal{O}(\log |E|)$}
            \State PQ.push($ |f(\tilde{w}_{ut}, \tilde{w}_{vt})|, e_{ut}$) \Comment{$\mathcal{O}(\log |E|)$  }
          \Else
            \State $\tilde{E} \gets \tilde{E} \cup \{e_{ut}\}$
            \State PQ.push($ |\tilde{w}_{vt}|, e_{ut}$) \Comment{$\mathcal{O}(\log |E|)$}
          \EndIf
        \EndFor
        \State
        \Return PQ, \texttt{canBeMerged}, $\tilde{E}$
      \EndFunction
    \end{algorithmic}
    \label{detailed_alg}
  \end{algorithm*}
  \begin{algorithm*}[p]
    \caption{Mutex Watershed Algorithm proposed by \cite{wolf2018mutex}}
  \hspace*{\algorithmicindent} \textbf{Input:} $\mathcal{G}(V,E,w^+,w^-)$ with $N$ nodes and $M$ edges \\
  \hspace*{\algorithmicindent} \textbf{Output:} Final clustering \\
    \hspace*{\algorithmicindent} 
    \begin{algorithmic}[1]
        \State \texttt{UF} $\gets$ initUnionFind($V$) 
        \For{$(u,v)=e\in E$ in descending order of $|w_e|=|w_e^+ - w_e^-|$}
          \If{\texttt{UF}.find($u$) $\neq$ \texttt{UF}.find($v$)} \Comment{Check if $u,v$ are already in the same cluster}
            \If{({\color{green}\textbf{$w_e > 0$}}) \textbf{and} \texttt{canBeMerged}($u,v$)}  \Comment{Check for cannot-link constraints}
              \State \texttt{UF}.merge($u,v$) and inherit constraints of parent clusters
            \ElsIf{({\color{red}\textbf{$w_e \leq 0$}})}
              \State Add cannot-link constraints between parent clusters of $u,v$
            \EndIf
          \EndIf
        \EndFor
        \State
        \Return Final clustering given by union-find data structure \texttt{UF}
    \end{algorithmic}
    \label{alg:mutex_watershed}
  \end{algorithm*}
  \clearpage
}

\paragraph{Implementation} Our implementation of \algname{} is based on an union-find data structure and a heap allowing deletion of its elements. 
In Phases 2 and 3, \algname{} is equivalent to a standard hierarchical agglomerative clustering algorithm with complexity $\mathcal{O}(N^2 \log N)$. In Algorithm \ref{detailed_alg}, we show our implementation of phase 1, involving cannot-link constraints.
In phase 1, the algorithm starts with each node assigned to its own cluster and sorts all edges $e\in E$ in a heap/priority queue (PQ) by their absolute weight $|\cost_e|=|w_e^+ - w_e^-|$ in descending order, so that the most attractive and the most repulsive interactions are processed first. It then iteratively pops one edge $e_{uv}$ from PQ and, depending on the priority $\tilde{\cost}_{uv}$, does the following: in case of attractive interaction $\tilde{\cost}_{uv}>0$, provided that $e_{uv}$ was not flagged as a cannot-link constraint, merge the connected clusters, perform an edge contraction of $e_{uv}$ in $\tilde{\mathcal{G}}_\Pi$ and update the priorities of new double edges as explained in Fig. \ref{fig:edge_contraction_and_contr_graph}. 
If, on the other hand, the interaction is repulsive ($\tilde{\cost}_{uv}\leq 0$) and the option \texttt{addCannotLinkContraints} of Alg. \ref{detailed_alg} is \texttt{True}, then the edge $e_{uv}$ is flagged as cannot-link constraint.

\begin{table}[t]
\centering
    \small
\begin{tabular}[t]{r | l }
            \toprule
            Linkage criteria & Update rule $f$ \\        
            \midrule
            Sum: & \thead[l]{$f(\tilde{\cost}_1,\tilde{\cost}_2) = \tilde{\cost}_1+\tilde{\cost}_2$} \\ 
            \makecell[r]{Absolute \\Maximum:} & \thead[l]{
            $
            f(\tilde{\cost}_1,\tilde{\cost}_2) = \begin{cases} 
            \tilde{\cost}_1 & \text{if}\,\, |\tilde{\cost}_1|>|\tilde{\cost}_2|\\
            \tilde{\cost}_2 & \text{otherwise}
             \end{cases} 
            $}
               \\ 
            \makecell[r]{Average:} & \thead[l]{$f(\tilde{\cost}_1,\tilde{\cost}_2) = \mathrm{weightAvg}\{ \tilde{\cost}_1, \tilde{\cost}_2 \} $}                 \\ 
            Single: & \thead[l]{$f(\tilde{\cost}_1,\tilde{\cost}_2) = \max \{ \tilde{\cost}_1, \tilde{\cost}_2 \}  $} \\
            Complete:& \thead[l]{$f(\tilde{\cost}_1,\tilde{\cost}_2) = \min \{ \tilde{\cost}_1, \tilde{\cost}_2 \}  $} 
        \end{tabular}\vspace{1em}
        \caption{The table lists the update rules $f(\tilde{\cost}_1, \tilde{\cost}_2)$ associated to the linkage criteria of Table \ref{tab:linkage-criteria} and that are used to efficiently update the interactions between clusters.}
\label{tab:linkage_criteria_explicit}  
\end{table}

\paragraph*{Complexity} 
In the main loop of Phase 1, the algorithm iterates over all edges, but the only iterations presenting a complexity different from $\mathcal{O}(1)$ are the ones involving a merge of two clusters, which are at most $N-1$. By using a union-find data structure (with path compression and union by rank) the time complexity of \texttt{merge}$(u, v)$ and \texttt{find}($u$) operations is $\mathcal{O}(\alpha(N))$, where $\alpha$ is the slowly growing inverse Ackerman function. The algorithm then iterates over the neighbors of the merged cluster (at most $N$) and updates/deletes values in the priority queue ($\mathcal{O}(\log |E|)$). 
Therefore, 
similarly to a heap-based implementation of hierarchical agglomerative clustering, our implementation of \algname{} - Phase 1 has a complexity of $\mathcal{O}(N^2 \log N)$. In the worst case, when the graph is dense and $|E|=N^2$, the algorithm requires $\mathcal{O}(N^2)$ memory. Nevertheless, in our practical applications the graph is much sparser, so $\mathcal{O}(|E|)=\mathcal{O}(N)$. 
With a single-linkage, corresponding to the choice of the \emph{Maximum} update rule in our framework, the algorithm can be implemented by using the more efficient Kruskal's Minimum Spanning Tree algorithm with complexity $\mathcal{O}(N \log N)$, but only when cannotLinkConstraints are not used. 
Moreover, \algname{} with \emph{Absolute Maximum} linkage can be implemented more efficiently (see next section). %

\begin{figure}
\includegraphics[width=\linewidth]{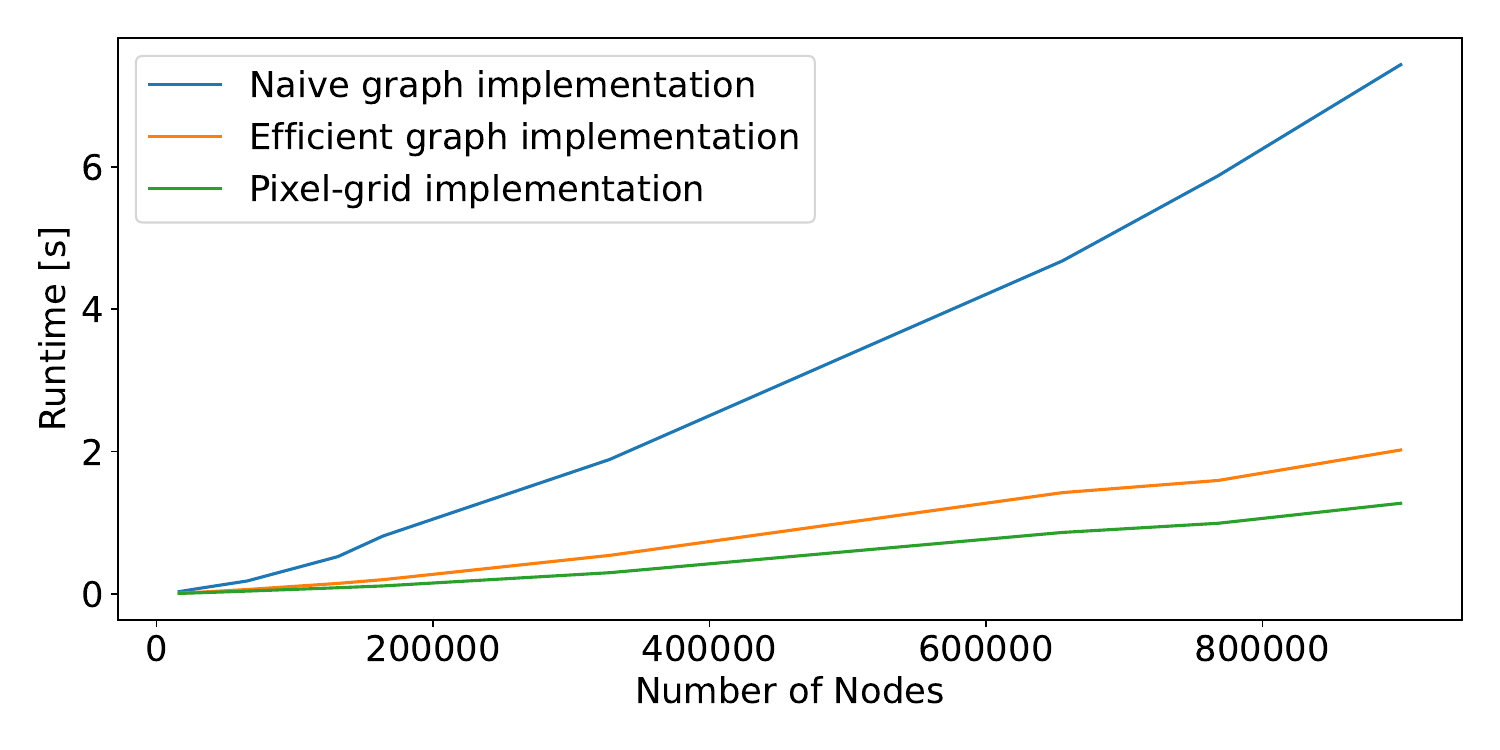}\\
\caption{Runtimes for different implementation of GASP with AbsMax linkage criterion. Runtimes are averaged over 5 runs.}
  \label{fig:runtimes_plot}
\end{figure}

\paragraph*{Efficiency of different GASP implementations with AbsMax linkage criteria} 
In Fig.~\ref{fig:runtimes_plot}, we compare the runtimes of three implementations of the AbsMax criteria: the implementation from \cite{wolf2019mutex} for pixel graphs (\emph{Pixel-grid implementation}) and for general graphs (\emph{Efficient graph implementation}) as well as the HC implementation with AbsMax linkage (\emph{Naive graph implementation}). The specialized implementations can exploit the properties of the underlying graph and are faster. But our generalization does not carry a large computational penalty and only requires a few extra seconds for partitioning graphs of a million nodes. Note that we have always used the most efficient implementation for the results reported in the paper. We will clarify this fact.

\paragraph{Median linkage} We implemented median linkage in our library from the beginning but did not report on it in the main paper for two reasons: we consider the other criteria to span the range of interesting behavior well; and it performs no better than some of the other criteria (like average linkage) which are faster to evaluate.

\begin{figure}
\includegraphics[width=\linewidth]{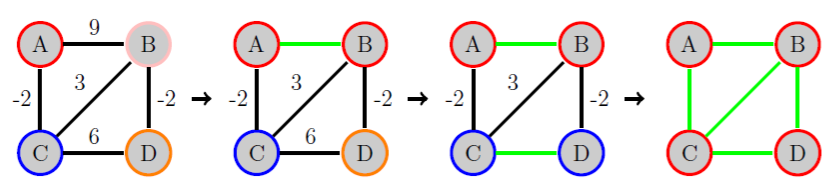}\\
\caption{GASP agglomeration with the Abs Max criterion: contracted edges are marked green. The last contraction increases the MC objective from -1 to 0.}\label{counterexample}
\end{figure}

\subsection{GASP relation to the multicut objective} \label{sec:relation_to_multicut}
For some of the linkage criteria, e.g. sum and average, GASP can be understood as a local search to the objective of the multicut optimization problem \ref{eq:MC_objective}, see \cite{levinkov2017comparative}. But this does not hold in general: the Abs Max linkage for example does not always decrease the MC objective (see counter example in Fig.~\ref{counterexample}). 
Moreover, GASP cannot be seen as a k-approximation, because it is a polynomial algorithm and \emph{Chawla, et al. Computational complexity, 2006}
has shown that approximating the multicut objective with any constant factor is in itself NP-hard.

\subsection{Proofs of Propositions \ref{prop:absmax_mutex}, \ref{prop:weight_shift_invariant}, \ref{prop:ultraMetric1}, and \ref{prop:ultraMetric2}}
\label{sec:proposition_proofs}

\begin{lemma} \label{lemma:absMax_and_complete_not_positive}
If \algname{} Algorithm \ref{main_alg} with \textbf{Complete linkage criteria} enforces a constraint between two clusters in \emph{Phase 1}, then the interaction between the clusters will never become positive over the course of the following agglomeration steps.
\end{lemma}
\begin{proof}
Two clusters are constrained in \emph{Phase 1} only if their interaction is repulsive and, with complete linkage,  the signed interaction between two clusters can only decrease over the course of the agglomeration. Thus, if two clusters are constrained by the algorithm, their negative interaction cannot increase and become positive later on in the agglomeration process.
\end{proof}

\begin{lemma} \label{lemma:absMax_and_complete_not_positive}
If \algname{} Algorithm \ref{main_alg} with \textbf{AbsMax linkage criteria} enforces a constraint between two clusters in \emph{Phase 1}, then the interaction between the clusters will never become positive over the course of the following agglomeration steps.
\end{lemma}
\begin{proof}
During the agglomeration the interaction between two clusters can only increase in absolute value. Thus, the negative interaction $\interact{}(S_i \cup S_j)<0$ between two constrained clusters can possibly become positive over the course of next agglomeration steps only if there is at least another pair of clusters in the graph that has a positive interaction $\interact{}(S_l \cup S_t)>0$ higher in absolute value: $|\interact{}(S_l \cup S_t)| > |\interact{}(S_i \cup S_j)|$.
If such clusters $S_l, S_t$ with positive interaction exist, we note that they must also be constrained (in the opposite case, the algorithm would have already merged them before to constrain $S_i$ and $S_j$, because their priority is higher). In other words, a constrained negative interaction can become positive only if there is already another positive constrained interaction: but this can never be the case because initially all constrained interactions are negative.
\end{proof}

\begin{lemma} \label{lemma:absMax_and_complete_property}
In the \algname{} Algorithm \ref{main_alg} with AbsMax or Complete linkage criteria (see linkage definition in Table \ref{tab:linkage-criteria}), the same final clustering is returned whether or not cannot-link constraints are enforced.
\end{lemma}
\begin{proof}
In phase 1 of Algorithm \ref{detailed_alg}, two clusters are merged only if the condition at line 9 is satisfied (i.e. when an interaction is both positive and not constrained). From Lemma \ref{lemma:absMax_and_complete_not_positive} and Lemma \ref{lemma:absMax_and_complete_not_positive} follows that with Complete and AbsMax linkage an interaction can never be both positive and constrained at the same time, so we directly conclude that the constrained and unconstrained versions of the algorithm will perform precisely the same agglomeration steps in phase 1.
In phase 2 (after constraints have been removed) no clusters are merged because all interactions are already negative (whether they previously constrained or not). Thus, both constrained and unconstrained versions of \algname{} return the same clustering $\Pi^*$.
\end{proof}

\absmaxmutex*

\begin{proof}%
From Lemma~\ref{lemma:absMax_and_complete_property} it directly follows that \algname{} with AbsMax linkage criterion returns the same final clustering whether or not cannot-link constraints are enforced. In the following, we prove that MWS (see pseudocode \ref{alg:mutex_watershed}) and the constrained AbsMax version of \algname{} also return the same clustering.
Both algorithms sort edges in descending order of the absolute interactions $|w_e|$ and then iterate over all of them. The only difference is that MWS, after merging two clusters, does not update the interactions between the new cluster and its neighbors. 
However, since with an Abs. Max. linkage the interaction between clusters is simply given by the edge with highest absolute weight $|w_e|$, the order by which edges are iterated over in \algname{} is never updated. Thus, both algorithms perform precisely the same steps and return the same clustering.
\end{proof}

\begin{figure}[t]
\centering
\includegraphics[width=0.48\textwidth,trim=0in 0in 0in 0in,clip]{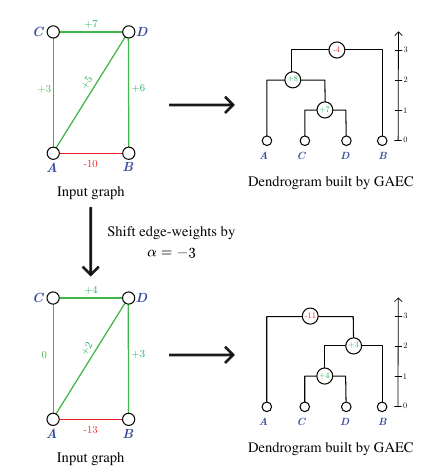}

        \caption{ Counter-example showing that GAEC is not weight-shift invariant.
        } \label{fig:counter_examples_shift_inv_GAEC}
\end{figure}
\begin{figure}[t!]
\centering
\includegraphics[width=0.48\textwidth,trim=0in 0in 0in 0in,clip]{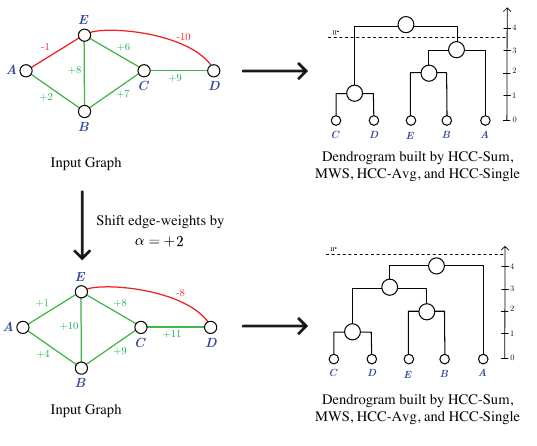}

        \caption{ Counter-example showing that HCC-Sum, MWS, HCC-Avg, and HCC-Single are not weight-shift invariant.
        } \label{fig:counter_examples_shift_inv}
\end{figure}

\invariantAlgs*
\begin{proof}
Theorem 1 in \cite{chehreghani2020hierarchical} proves that hierarchical clustering with Average (HC-Avg), Single (HC-Single), and Complete linkage (HC-Complete) are weight-shift invariant.

The same is not true for \algname{} with Sum linkage criteria (GAEC and HCC-Sum), because by adding a constant $\alpha$ to all edge weights $w_e$, the interaction between two clusters $S_i$ and $S_j$ is increased by a factor $\alpha |E_{ij}|$, which depends on the number of edges  $|E_{ij}|$ connecting the two clusters. Thus, when all edge weighs of the graph are shifted, the agglomeration order may change. For a simple example of this, it is enough to consider the toy graph in Fig.~\ref{fig:intro_figure}a and shift the weights of the graph by $\alpha=-3$ (see Fig.~\ref{fig:counter_examples_shift_inv_GAEC}).

The constrained versions of \algname{} (HCC-Avg and HCC-Single) are also not weight-shift invariant: here, the algorithm merges or constrains clusters in a given order, depending on the absolute interactions $|\interact{}(S_i \cup  S_j)|$ between clusters; so, when edge weights are shifted by a constant $\alpha$, the sorting by absolute value can change arbitrarily together with the agglomeration order, as we show in the counter-example of Fig.~\ref{fig:counter_examples_shift_inv}.
Similarly, the Mutex Watershed algorithm is not weight-shift invariant because it uses a linkage criterion that compares weights by their absolute values (see again counter-example in Fig.~\ref{fig:counter_examples_shift_inv}.
\end{proof}

\begin{restatable}{prop}{firstUltraMetricProperty}
\label{prop:ultraMetric1}
Consider a graph $\mathcal{G}(V,E,w_e)$, a linkage criterion $\interact{}$, and an agglomerative algorithm returning a binary rooted tree $T$ with height $h_T$. Then, $(V, d_{T})$ defined in Eq.~\ref{eq:UM_def} is an ultrametric if and only if the following is true:
\begin{gather}
\forall u,v,t \in V \nonumber \\
 h_T(u, v) < h_T(u, t) \Rightarrow \interact{}_{T}(u,v) \geq \interact{}_{T}(u,t) \label{eq:UM_assumption}
\end{gather}
In words, condition \ref{eq:UM_assumption} means: if the algorithm merges nodes $u,v$ before to merge nodes $u,t$, then the signed interaction $\interact{}_{T}(u,v)$ between $u$ and $v$ has to be higher or equal than $\interact{}_{T}(u,t)$.
\end{restatable}
\begin{proof}

From the definition of $d_{T}$, it follows that:
\begin{align}
d_{T}(u,u) &= 0 \qquad &\forall u\in V \label{eq:dist_1} \\ 
d_{T}(u,v) &\geq 0 \qquad &\forall u,v \in V \label{eq:dist_2}\\
d_{T}(u,v)& =d_{T}(v,u) \qquad &\forall u,v \in V. \label{eq:dist_3}
\end{align}
In order to show that $(V, d_{T})$ is an ultrametric, we only need to prove the ultrametric property:
\begin{equation}\label{eq:UM_property_original}
d_T(u,v) \leq \max \{d_T(u,t), d_T(v,t)\} \quad \forall u,v,t \in V.
\end{equation}
When at least two of the three nodes $u,v,t \in V$ are the same, this property follows from Eq.~\ref{eq:dist_1} and Eq.~\ref{eq:dist_2}. When nodes $u,v,t\in V$ are distinct, from the definition of $d_{T}$ it follows that Eq.~\ref{eq:UM_property_original} is equivalent to:
\begin{equation}\label{eq:UM_property}
\interact{}_{T}(u,v) \geq \min \{\interact{}_{T}(u,t), \interact{}_{T}(v,t)\}.
\end{equation}
In the following, we prove both sides of the \emph{if and only if} statement in the proposition. First, we prove the $(\Leftarrow)$ side, i.e. that if assumption \ref{eq:UM_assumption} holds, then $(V, d_{T})$ is an ultrametric and \ref{eq:UM_property} holds. 

Case 1: in Eq.~\ref{eq:UM_property}, $t\in V$ is part of the sub-tree ${T[u \vee v]}$. In other words, the algorithm first merges node $t$ with either node $u$ or $v$, and then $u$ and $v$ are merged together. Let us assume that $t$ is first merged with $u$ (the following proof also holds for the opposite case in which $t$ is first merged with $v$):
\begin{equation}\label{eq:case_1}
h_T(u, t) < h_T(u, v) = h_T(v, t).
\end{equation}
Thus, by combining the last equation with assumption (\ref{eq:UM_assumption}), it follows that
\begin{equation}
\interact{}_{T}(u, t) \geq \interact{}_{T}(v, t) \quad  \text{and} \quad \interact{}_{T}(u, v) = \interact{}_{T}(v, t)
\end{equation}
and Eq.~\ref{eq:UM_property} follows (becoming an equality in this case).

Case 2: in Eq.~\ref{eq:UM_property}, $t\in V$ is \emph{not} part of the sub-tree ${T[u \vee v]}$. Thus, the algorithm first merges nodes $u$ and $v$, and then it merges node $t$ together with the cluster containing $u$ and $v$:
\begin{equation}
h_T(u, v) < h_T(u, t) = h_T(v, t).
\end{equation}
Thus, from assumption \ref{eq:UM_assumption} we have that
\begin{equation}
\interact{}_{T}(u, v) \geq \interact{}_{T}(u, t) \quad  \text{and} \quad \interact{}_{T}(u, v) \geq \interact{}_{T}(v, t),
\end{equation}
so also in this case Eq.~\ref{eq:UM_property} follows.

Next, we are left to prove the $(\Rightarrow)$ side of the \emph{if and only if} statement: if $(V, d_{T})$ is an ultrametric, then assumption \ref{eq:UM_assumption} holds.
To prove this statement, we first rephrase it in the following equivalent form: if assumption \ref{eq:UM_assumption} does not hold, then $(V, d_{T})$ is not an ultrametric and \ref{eq:UM_property} does not hold. If we negate assumption \ref{eq:UM_assumption}, there must be at least three $u,v,t \in V$ such that: 
\begin{equation}
h_T(u, v) < h_T(u, t) \quad \text{and} \quad  \interact{}_{T}(u,v) < \interact{}_{T}(u,t).
\end{equation}
The first condition, in words, is again assuming that the algorithm first merges nodes $u$ and $v$, and later it also merges node $t$ with the cluster containing $u$ and $v$. Thus, we can rephrase this assumption as:
\begin{equation}
\interact{}_{T}(u, v) < \interact{}_{T}(u, t) = \interact{}_{T}(v, t).
\end{equation}
From this, it follows that
\begin{equation}
\interact{}_{T}(u, v) < \min \{\interact{}_{T}(u, t), \interact{}_{T}(v, t)\},
\end{equation}
which is exactly the negation of the ultrametric property \ref{eq:UM_property}.
\end{proof}

\secondUltraMetricProperty*
\begin{proof}
Thanks to Prop.~\ref{prop:ultraMetric1}, we know that $(V, d_{T^*})$ is an ultrametric if and only if assumption \ref{eq:UM_assumption} holds. Thus, in the following, we will prove which variations of the \algname{} Algorithm \ref{main_alg} satisfy assumption \ref{eq:UM_assumption}. 
In other words, we need to prove in which cases \algname{} merges clusters according to a monotonously decreasing order of signed interactions $\interact{}$.

\algname{} puts clusters in a priority queue (Algorithm \ref{main_alg}, lines 5 and 15) and merges them starting from those with the highest interaction (lines 9, 19, and 26). However, the priority queue is updated each time two clusters are merged (lines 10, 20, and 27). Thus, to ensure a monotonously decreasing merging order, updated interactions involving a merged cluster should always be lower or equal than previously existing interactions (\textbf{condition 1}):
\begin{align}\label{eq:condition1}
& \forall S_i \in \Pi \setminus \{S_1, S_2\}, \nonumber \\
\interact{}(S_1 \cup S_2 \cup  S_i) &\leq \max \{ \interact{}(S_1  \cup  S_i), \interact{}(S_2  \cup  S_i)\} 
\end{align}
where $\Pi$ is a clustering, $\interact{}$ is a linkage criteria, and $S_1,S_2\in \Pi$ are two clusters merged by the algorithm at a given iteration. If this condition is true then, in the following iterations, \algname{} can only merge clusters with lower (or equal) interaction values. 

We also note that, in phase 1, the algorithm skips interactions that are both positive and constraint (condition at line 8 in Algorithm \ref{main_alg}) and merges them only later in phase 2 (line 19), when constraints are removed.
Clearly, whenever this happens, a decreasing merging order is no longer ensured. 
Thus, on top of condition 1, we also have that no merging decisions should be ``delayed'' from phase 1 to phase 2 (\textbf{condition 2}). 

Condition 1 always holds for Average, Single, Complete, and AbsMax linkage criteria, but not for a Sum linkage criteria, because the sum of two positive numbers $a,b$ is always higher than $\max\{a,b\}$. This is also demonstrated in the toy example of Fig.~\hyperref[fig:intro_figure]{\ref*{fig:intro_figure}a}, proving that, in general, Sum-linkage algorithms like GAEC or HCC-Sum do not define an ultrametric on the graph.

Thanks to Lemma \ref{lemma:absMax_and_complete_property}, we have that condition 2 always holds for algorithms based on AbsMax and Complete linkage, proving that the Mutex Watershed and HC-Complete algorithms define an ultra-metric (whether or not cannot-link-constraints are enforced). On the other hand, condition 2 does not hold for other variations of \algname{} involving cannot-link-constraints (HCC-Sum, HCC-Avg, and HCC-Single), which do not then define an ultrametric. 

Finally, the remaining not constrained versions of \algname{} (HC-Avg, HC-Single, and HC-Complete) satisfy both conditions, so they define an ultrametric, confirming the well-known results of related work in hierarchical clustering on unsigned graphs \cite{johnson1967hierarchical,milligan1979ultrametric}.

\end{proof}

\subsection{Mutex Watershed on SSBM graphs}
\begin{prop} \label{prop:MWS_on_SSBM}
Consider a graph generated by an Erd\H os-R\'enyi signed stochastic block model (SSBM) as described in Section \ref{sec:clustering_problems}, with $N$ nodes, edges added with probability $p$, sign-flip probability $\eta<0.5$, $k$ ground-truth clusters, and edge weights Gaussian-distributed with standard deviation $\sigma$. Then, at every iteration, \algname{} with Absolute Maximum linkage (or, in other words, the Mutex Watershed algorithm) always makes a mistake with at least probability $\eta$. 
\end{prop}
\begin{proof}
Thanks to Lemma \ref{lemma:absMax_and_complete_property} we know that \algname{} with Absolute Maximum linkage returns the same clustering whether or not cannot-link-constraints are used. Thus, in the following, we prove the proposition considering the version enforcing constraints.
Let us consider a generic iteration of the algorithm, where two clusters $S_\alpha$ and $S_\beta$ have the highest priority and are popped from priority queue. Then, the MWS algorithm will either merge or constrain them depending on the fact that their interaction $\interact_{\mathrm{AbsMax}}(S_\alpha  \cup S_\beta)$ is  positive or negative (note that, with AbsMax linkage, an interaction can never be positive and constrained, as shown in Lemma \ref{lemma:absMax_and_complete_property}).
By construction of the SSBM, every edge $e\in E$ in the graph has a absolute weight distributed as $|w_e|\sim \mathcal{N}(1,\sigma^2)$. Thus, every edge $e'\in(S_{\alpha} \times S_{\beta})\cap E$ connecting the two clusters has the same probability to have the highest absolute weight, and the sign of the interaction $\interact_{\mathrm{AbsMax}}(S_\alpha  \cup  S_\beta)$ will only depend on the sign of this highest edge. Therefore, the probability that the MWS merges two clusters is simply given by the fraction of positive weighted edges connecting them.

Let $\tilde{\Pi}=\{\tilde{S}_1,\ldots,\tilde{S}_k\}$ denote the ground truth clustering, and $\tilde{S}_{\alpha i}=S_\alpha\cap \tilde{S}_i$ denote the intersection between cluster $S_{\alpha}$ and a ground-truth cluster $\tilde{S}_i$.
If the generated graph is dense, i.e. $p=1$, then the total number of edges connecting clusters $S_{\alpha}$ and $S_{\beta}$ that have a true attractive or repulsive weight is (according to the ground truth labels)
\begin{equation}
\NBE^+=\sum_{i=1}^k |\tilde{S}_{\alpha i}||\tilde{S}_{\beta i}|, \quad \NBE^-=\sum_{i=1}^k \sum_{j=1,j\neq i}^k |\tilde{S}_{\alpha i}||\tilde{S}_{\beta j}|.
\end{equation}
When the edges in the graph are randomly added with a probability $p$, then the actual number of true attractive and repulsive interactions connecting the two clusters is (according to the ground truth labels):
\begin{equation}
\nBE^+ \sim \mathcal{B}(\NBE^+,p), \qquad \nBE^- \sim \mathcal{B}(\NBE^-,p), 
\end{equation}
where $\mathcal{B}(\NBE,p)$ is the binomial distribution:
\begin{equation}
\mathcal{B}(\nBE;\NBE,p) = \frac{\NBE !}{\nBE! (\NBE-\nBE)!} p^\nBE (1-p)^{\NBE-\nBE}.
\end{equation}
Here, we only assume that $\nBE^+ + \nBE^- >0$, i.e. there is at least one edge connecting the two clusters (otherwise their interaction would be zero and the MWS would not have popped them from priority queue). 

So far we have been talking about attractive and repulsive connections according to the ground truth labels. In our SSBM however every edge has a uniform probability $\eta$ to have its sign flip, so  the actual number of attractive interactions connecting the two clusters will be instead given by the sum of the true attractive interactions $\nBE^+_{\mathrm{nf}}\sim \mathcal{B}(\nBE^+,1-\eta)$ that have not been flipped, plus the true negative interactions $\nBE^-_{\mathrm{f}}\sim \mathcal{B}(\nBE^-,\eta)$ that have been flipped.
Putting everything together, given two clusters with $\nBE^+$ true attractive interactions and $\nBE^-$ true negative ones, the highest-absolute-weight edge connecting them has the following probability to be positive:
\begin{align}\label{eq:mws_prob}
 \mathbb{P}[\interact_{\mathrm{AbsMax}}&(S_\alpha  \cup  S_\beta)>0;\nBE^+,\nBE^-] = \nonumber \\
=&\sum_{\nBE^+_{\mathrm{nf}}=0}^{\nBE^+}\sum_{\nBE^-_{\mathrm{f}}=0}^{\nBE^-} 
\mathcal{B}(\nBE^-_{\mathrm{f}};\nBE^-,\eta) \mathcal{B}(\nBE^+_{\mathrm{nf}};\nBE^+,1-\eta)\cdot  \nonumber\\
& \qquad \qquad  \cdot \left(\frac{\nBE^+_{\mathrm{nf}}+\nBE^-_{\mathrm{f}}}{\nBE^+ + \nBE^-} \right) \nonumber\\
\stackrel{(*)}{=}&\frac{\nBE^+(1-\eta)+\nBE^-\eta}{\nBE^+ + \nBE^-}
\end{align}  
where in $(*)$ we used the fact that the expected value of a binomial distribution $\mathcal{B}(\nBE,\eta)$ is $\nBE \eta$.

Now we note that this probability is bounded in the interval $[\eta, 1-\eta]$. So, regardless of whether the two clusters $S_\alpha$ and $S_\beta$ should be merged or constraint according to ground truth labels, the probability not to make the correct decision is always at least $\eta$.
Remarkably, while the exact probability in Eq.~\ref{eq:mws_prob} depends on the number of edges connecting the two clusters $\nBE^++\nBE^-$ and thus on the cluster sizes, the bounds do not. Thus, this result shows that, unlike Sum or Avg linkage methods, the MWS algorithm is unable to reliably correct for the sign flip noise even for big clusters linked by many edges.
\end{proof}

\subsection{Application to neuron segmentation}\label{sec:cremi_details}
\paragraph{Training and data augmentation} The data from the CREMI challenge is highly anisotropic and contains artifacts like missing sections, staining precipitations and support film folds. 
To alleviate difficulties stemming from misalignment, we use a version of the data that was elastically realigned by the challenge organizers with the method of \emph{S.    Saalfeld, et al. Nature methods, 2012}.
In addition to the standard data augmentation techniques of random rotations, random flips and  elastic deformations, we simulate data artifacts.
We randomly zero-out slices, decrease the contrast of slices, simulate tears, introduce alignment jitter and paste artifacts extracted from the training data. Both \cite{funke2018large} and \cite{lee2017superhuman} have shown
that these kinds of augmentations can help to alleviate issues caused by EM-imaging artifacts.
We use L2 loss and Adam optimizer to train the network. The model was trained on all three samples with available ground truth labels.  

\paragraph{CREMI-gridRag instances} 
Our 3D UNet model predicts the same set of 12 long-and-short range affinities as described in \cite{lee2017superhuman}. 
When building the pixel-grid graph, we add both direct neighbors connections and the long-range connections predicted by our model (every voxel is connected to other six voxels via direct connections and other 18 voxels via long-range edges).
Empirically, when long-range predictions of the CNN are added as long-range connections in the graph, \algname{} achieves better scores as compared to when only direct-neighbors predictions are used.
Our intuitive explanation of this is that, where there is a clear boundary evidence between two segments, the long-range predictions of the CNN model are more certain than the direct-neighbor ones, because it is often impossible to estimate the exact ground-truth label transition for pixels that are very close to a boundary evidence. 
However, empirically, we also find that \algname{} achieves the best scores when only 10\% of the long-range connections are randomly sampled and added to the grid-graph. When all the long-range connections predicted by the CNN are added to the graph (18 connections for every voxel), all versions of \algname{} tend to perform more over-clustering errors.
In practice, we explain this by observing that many challenging parts of the studied neuron segmentation data involve thin and elongated segments, and our model sometimes fails to connect distant pairs of pixels that, according to the ground-truth labels, should belong to the same segment (even though, in this case, the direct neighboring predictions are correct).
To sum up, the scores we report in Tables \ref{tab:scores_gridGraph} are obtained by using only 10\% of the long-range predictions, since this was the setup that performed the best.
After running \algname{}, we use a simple post-processing step to delete small segments on the boundaries, most of which are given by single-voxel clusters. On the neuron segmentation predictions, we deleted all regions with less than 200 voxels and used a seeded watershed algorithm to expand the bigger segments.

\paragraph{CREMI-3D-rag instances} 
We build these clustering problems by generating superpixels and then building a 3D region adjacency graph.
Due to the anisotropy of the data, we generate 2D superpixels by considering each 2D image in the stack singularly.
First, we generate a boundary-evidence map by taking an average over the two direct-neighbor predictions of the CNN model (one for each direction in the 2D image of the stack) and applying some additional smoothing. Then, we threshold the boundary map, compute a distance transform, and run a watershed algorithm seeded at the maxima of the distance transform (WSDT). The degree of smoothing was optimized such that each region receives as few seeds as possible, without however causing severe under-segmentation. 
The computed 2D superpixels are then used to build a 3D region-adjacency graph (3D-rag). The weights of the edges are given by averaging the CNN affinities over the boundaries of adjacent superpixels.

\begin{table*}[t]
    \centering
    \scriptsize
    \begin{subtable}[t!]{\textwidth}
    \centering

        \begin{tabular}{l | r r r r r r r r r}
        Clustering problem & \multicolumn{1}{r}{GAEC \cite{keuper2015efficient}} & HCC-Sum & MWS \cite{wolf2018mutex} & HC-Avg & HCC-Avg & HC-Single & HCC-Single & HC-Complete \\ \midrule
        \emph{Modularity Clustering} & 
        -0.457 & -0.453 & -0.073 & \textbf{-0.467} & \textbf{-0.467} & 0.000 & 0.000 & -0.201 \\ 
        \emph{Image Segmentation}  & 
        \textbf{-2,955} & -2,953 & -2,901 & -2,903 & -2,896 & -1,384 & -1,384 & -2,102 \\
        \emph{Knott-3D (150-300-450)}  & 
        \textbf{-36,667} & -36,652 & -35,200 & -35,957 & -35,631 & -2,522 & -2,522 & 30,629 \\
        \emph{CREMI-3D-rag}  
        & \textbf{-1,112,287} & -1,112,286& -1,109,731 & -1,112,177 & -1,112,100 & -1,038,709 & -1,038,709 & -748,734,869 \\ 
        \emph{Fruit-Fly Level 1-4}
        & \textbf{-151,022} & -151,017 & -150,879 & -150,909 & -150,876 & -71,477 & -71,997 & -128,733 \\
        \emph{CREMI-gridGraph} 
        & -73,317,601 & -73,328,867 & -73,330,568 & \textbf{-73,502,947} & -73,474,856 & -45,194,180 &-45,194,443 & 311,598,700 \\
        \emph{Fruit-Fly Level Global} 
        & \textbf{-151,688} & -151,596 & -146,315 & -150,466 & -150,171 & -4,422 & - & 6,876 \\

        \end{tabular}
    \end{subtable} 
    \caption{We compare algorithms in the \algname{} framework by evaluating which of the obtained clusterings is associated to the lowest value of the multicut objective defined in Eq.~\ref{eq:MC_objective} (lower is better). Single and complete linkage methods performed much worse than the others. Note that HCC-Single is the algorithm with the highest runtime (see Table \ref{tab:scores_gridGraph}) and it did not scale up to the very large clustering problem \emph{Fruit-Fly Level Global}.
    } 
    \label{tab:all_multicut_energies}
\end{table*}

\subsection{Adding structured noise to CNN predictions} \label{sec:appendix_noise_gen}
Additionally to the comparison on the full training dataset, we performed more experiments on a crop of the more challenging CREMI training sample B, where we perturbed the predictions of the CNN with noise and we introduced additional artifacts like missing boundary evidences.

\begin{figure*}[t]
\centering
        \includegraphics[width=\textwidth,trim=0.0in -0.in -0.0in -0.4in,clip]{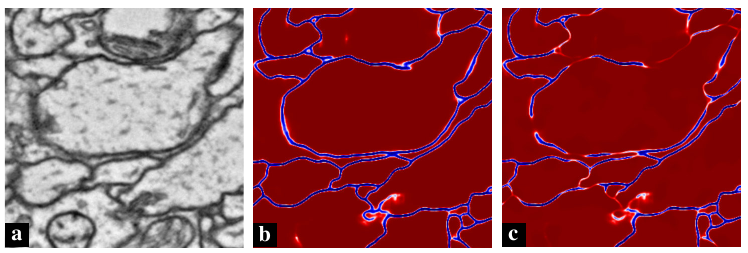}
    \caption{CNN predictions on a slice of the CREMI neuron segmentation challenge with and without additional spatially-correlated noise. (\textbf{a}) Raw data (\textbf{b}) Original CNN predictions $F(x)$, where blue pixels represent boundary evidence (\textbf{c}) Strongly perturbed version $\tilde{F}(x;\mathcal{K})$ of the predictions defined in Eq. \ref{eq:noise_biased_predictions} with $\mathcal{K}=8$. Long-range predictions are not shown. 
    }
    \label{fig:noisy_affs}
\end{figure*}

In the field of image processing there are several ways of adding noise to an image, among which the most common are Gaussian noise or Poisson shot noise. 
In these cases, the noise of one pixel does not correlate with its neighboring noise values. On the other hand, predictions of a CNN are known to be spatially correlated. 
Thus, we used Perlin noise\footnote{In our experiments, we used an open-source implementation of simplex noise \cite{perlin2001noise}, which is an improved version of Perlin noise \cite{perlin1985image}}, one of the most common gradient noises used in procedural pattern generation. This type of noise $n(x)\in[0,1]$ generates spatial random patterns that are locally smooth but have large and diverse variations on bigger scales. We then combined it with the CNN predictions $p(x)$ in the following way:
\begin{equation}\label{eq:noise_biased_predictions}
\tilde{F}(x;\mathcal{K})=F(x)+\mathcal{K}\cdot\max\left(N(x),0\right),
\end{equation}
where  $N(x)=\mathrm{Logit}[n(x)]$; $F(x)=\mathrm{Logit}[p(x)]$ and $\mathcal{K}\in \mathbb{R}^+$ is a positive factor representing the amount of added noise. The resulting perturbed predictions $\tilde{F}(x;\mathcal{K})$ are then under-clustering biased, such that the probability for two pixels to be in the same cluster is increased only if $N(x)>0$ (see Fig.~\hyperref[fig:noisy_affs]{\ref*{fig:noisy_affs}b} and \hyperref[fig:noisy_affs]{\ref*{fig:noisy_affs}c}). 
Note that in these experiments we focused only on predictions perturbed with under-clustering biased noise (and not over-clustering biased noise). The reason is that generating realistic over-clustering biased CNN predictions is more complex and cannot be simply done by adding Perlin noise: as we show in Fig.~\hyperref[fig:noisy_affs]{\ref*{fig:noisy_affs}c}, by adding Perlin noise we can easily ``remove'' parts of a boundary evidence, but it is not possible to generate random new realistic boundary evidence.  

In our experiments, each pixel is represented by a node in the grid-graph and it is linked to $n_{\mathrm{nb}}$ other nodes by short- and long-range edges. Thus, the output volume of our CNN model is a four-dimensional tensor with $n_{\mathrm{nb}}$ channels: for each pixel / voxel, the model outputs $n_{\mathrm{nb}}$ values representing affinities of different edge connections. We then generated a 4-dimensional Perlin noise tensor that matches the dimension of the CNN output. The data is highly anisotropic, i.e. it has a lower resolution in one of the dimensions. Due to this fact, we chose different smoothing parameters to generate the noise in different directions. 

\begin{table*}[t]
\centering
\begin{tabular}[t]{l c}
           Method & ARAND Error \\ \midrule
           \textbf{HC-Avg} (GASP with Avg Linkage) & \textbf{0.1034} \\
GAEC \cite{keuper2015efficient} (GASP with Sum Linkage) & 0.1035 \\
MWS \cite{wolf2018mutex} (GASP with AbsMax linkage) & 0.1068 \\
SPONGE$_{sym}$ \cite{Cucuringu2019SPONGEAG} & 0.4161\\
$L_{sym}$ \cite{kunegis2010spectral} & 0.8069 \\
SPONGE \cite{Cucuringu2019SPONGEAG} & 0.9211 \\
BNC \cite{chiang2012scalable} & 0.9926 \\
        \end{tabular}
    \caption{\algname{} compared to spectral clustering methods on a small crop of the CREMI neuron segmentation dataset. 
    Since spectral methods cannot scale to the full CREMI dataset, we evaluated them on a smaller $10\times100\times100$ sub-volume of CREMI training sample B.
    Despite the fact that the true number of ground truth clusters was given as an input to the spectral methods, GASP significantly outperformed them. 
    }
    \label{tab:cremi_spectral_experiments}
\end{table*}

%% file: agglo_clust_review.bbl
\begin{thebibliography}{10}\itemsep=-1pt

\bibitem{andres2011probabilistic}
Bjoern Andres, J{\"o}rg~H Kappes, Thorsten Beier, Ullrich K{\"o}the, and Fred~A
  Hamprecht.
\newblock Probabilistic image segmentation with closedness constraints.
\newblock In {\em 2011 International Conference on Computer Vision}, pages
  2611--2618. IEEE, 2011.

\bibitem{andres2012globally}
Bjoern Andres, Thorben Kroeger, Kevin~L Briggman, Winfried Denk, Natalya
  Korogod, Graham Knott, Ullrich Koethe, and Fred~A Hamprecht.
\newblock Globally optimal closed-surface segmentation for connectomics.
\newblock In {\em European Conference on Computer Vision}, pages 778--791.
  Springer, 2012.

\bibitem{arbelaez2011contour}
Pablo Arbelaez, Michael Maire, Charless Fowlkes, and Jitendra Malik.
\newblock Contour detection and hierarchical image segmentation.
\newblock {\em IEEE transactions on pattern analysis and machine intelligence},
  33(5):898--916, 2011.

\bibitem{arganda2015crowdsourcing}
Ignacio Arganda-Carreras, Srinivas~C Turaga, Daniel~R Berger, Dan
  Cire{\c{s}}an, Alessandro Giusti, Luca~M Gambardella, J{\"u}rgen Schmidhuber,
  Dmitry Laptev, Sarvesh Dwivedi, Joachim~M Buhmann, et~al.
\newblock Crowdsourcing the creation of image segmentation algorithms for
  connectomics.
\newblock {\em Frontiers in neuroanatomy}, 9:142, 2015.

\bibitem{bai2017deep}
Min Bai and Raquel Urtasun.
\newblock Deep watershed transform for instance segmentation.
\newblock In {\em Proceedings of the IEEE Conference on Computer Vision and
  Pattern Recognition}, pages 5221--5229, 2017.

\bibitem{bailoni2020proposal}
Alberto Bailoni, Constantin Pape, Steffen Wolf, Anna Kreshuk, and Fred~A
  Hamprecht.
\newblock Proposal-free volumetric instance segmentation from latent
  single-instance masks.
\newblock {\em arXiv preprint arXiv:2009.04998}, 2020.

\bibitem{bansal2004correlation}
Nikhil Bansal, Avrim Blum, and Shuchi Chawla.
\newblock Correlation clustering.
\newblock {\em Machine learning}, 56(1-3):89--113, 2004.

\bibitem{beier2016efficient}
Thorsten Beier, Bj{\"o}rn Andres, Ullrich K{\"o}the, and Fred~A Hamprecht.
\newblock An efficient fusion move algorithm for the minimum cost lifted
  multicut problem.
\newblock In {\em European Conference on Computer Vision}, pages 715--730.
  Springer, 2016.

\bibitem{beier2014cut}
Thorsten Beier, Thorben Kroeger, Jorg~H Kappes, Ullrich Kothe, and Fred~A
  Hamprecht.
\newblock Cut, glue \& cut: A fast, approximate solver for multicut
  partitioning.
\newblock In {\em Proceedings of the IEEE Conference on Computer Vision and
  Pattern Recognition}, pages 73--80, 2014.

\bibitem{beier2017multicut}
Thorsten Beier, Constantin Pape, Nasim Rahaman, Timo Prange, Stuart Berg,
  Davi~D Bock, Albert Cardona, Graham~W Knott, Stephen~M Plaza, Louis~K
  Scheffer, et~al.
\newblock Multicut brings automated neurite segmentation closer to human
  performance.
\newblock {\em Nature Methods}, 14(2):101, 2017.

\bibitem{brandes2007modularity}
Ulrik Brandes, Daniel Delling, Marco Gaertler, Robert Gorke, Martin Hoefer,
  Zoran Nikoloski, and Dorothea Wagner.
\newblock On modularity clustering.
\newblock {\em IEEE transactions on knowledge and data engineering},
  20(2):172--188, 2007.

\bibitem{chehreghani2020hierarchical}
Morteza~Haghir Chehreghani.
\newblock Hierarchical correlation clustering and tree preserving embedding.
\newblock {\em arXiv preprint arXiv:2002.07756}, 2020.

\bibitem{cheng2019panopticdeeplab}
Bowen Cheng, Maxwell~D. Collins, Yukun Zhu, Ting Liu, Thomas~S. Huang, Hartwig
  Adam, and Liang-Chieh Chen.
\newblock Panoptic-{DeepLab}.
\newblock {\em arXiv preprint arXiv:1910.04751}, 2019.

\bibitem{chiang2012scalable}
Kai-Yang Chiang, Joyce~Jiyoung Whang, and Inderjit~S Dhillon.
\newblock Scalable clustering of signed networks using balance normalized cut.
\newblock In {\em Proceedings of the 21st ACM international conference on
  Information and knowledge management}, pages 615--624. ACM, 2012.

\bibitem{chopra1991multiway}
Sunil Chopra and Mendu~R Rao.
\newblock On the multiway cut polyhedron.
\newblock {\em Networks}, 21(1):51--89, 1991.

\bibitem{chopra1993partition}
Sunil Chopra and Mendu~R Rao.
\newblock The partition problem.
\newblock {\em Mathematical Programming}, 59(1-3):87--115, 1993.

\bibitem{cciccek20163d}
{\"O}zg{\"u}n {\c{C}}i{\c{c}}ek, Ahmed Abdulkadir, Soeren~S Lienkamp, Thomas
  Brox, and Olaf Ronneberger.
\newblock {3D U-Net}: learning dense volumetric segmentation from sparse
  annotation.
\newblock In {\em International conference on medical image computing and
  computer-assisted intervention}, pages 424--432. Springer, 2016.

\bibitem{ciresan2012deep}
Dan Ciresan, Alessandro Giusti, Luca~M Gambardella, and J{\"u}rgen Schmidhuber.
\newblock Deep neural networks segment neuronal membranes in electron
  microscopy images.
\newblock In {\em Advances in neural information processing systems}, pages
  2843--2851, 2012.

\bibitem{Cucuringu2019SPONGEAG}
Mihai Cucuringu, Peter Davies, Aldo Glielmo, and Hemant Tyagi.
\newblock {SPONGE}: A generalized eigenproblem for clustering signed networks.
\newblock In {\em AISTATS}, 2019.

\bibitem{cucuringu2016simple}
Mihai Cucuringu, Ioannis Koutis, Sanjay Chawla, Gary Miller, and Richard Peng.
\newblock Simple and scalable constrained clustering: a generalized spectral
  method.
\newblock In {\em Artificial Intelligence and Statistics}, pages 445--454,
  2016.

\bibitem{de2017semantic}
Bert De~Brabandere, Davy Neven, and Luc Van~Gool.
\newblock Semantic instance segmentation with a discriminative loss function.
\newblock {\em arXiv preprint arXiv:1708.02551}, 2017.

\bibitem{demaine2006correlation}
Erik~D Demaine, Dotan Emanuel, Amos Fiat, and Nicole Immorlica.
\newblock Correlation clustering in general weighted graphs.
\newblock {\em Theoretical Computer Science}, 361(2-3):172--187, 2006.

\bibitem{fathi2017semantic}
Alireza Fathi, Zbigniew Wojna, Vivek Rathod, Peng Wang, Hyun~Oh Song, Sergio
  Guadarrama, and Kevin~P Murphy.
\newblock Semantic instance segmentation via deep metric learning.
\newblock {\em arXiv preprint arXiv:1703.10277}, 2017.

\bibitem{felzenszwalb2004efficient}
Pedro~F Felzenszwalb and Daniel~P Huttenlocher.
\newblock Efficient graph-based image segmentation.
\newblock {\em International journal of computer vision}, 59(2):167--181, 2004.

\bibitem{finkel2008enforcing}
Jenny~Rose Finkel and Christopher~D Manning.
\newblock Enforcing transitivity in coreference resolution.
\newblock In {\em Proceedings of the 46th Annual Meeting of the Association for
  Computational Linguistics on Human Language Technologies: Short Papers},
  pages 45--48. Association for Computational Linguistics, 2008.

\bibitem{funke2015learning}
Jan Funke, Fred~A Hamprecht, and Chong Zhang.
\newblock Learning to segment: training hierarchical segmentation under a
  topological loss.
\newblock In {\em International Conference on Medical Image Computing and
  Computer-Assisted Intervention}, pages 268--275. Springer, 2015.

\bibitem{cremiChallenge}
Jan Funke, Stephan Saalfeld, Davi Bock, Srini Turaga, and Eric Perlman.
\newblock {Cremi challenge}.
\newblock \url{https://cremi.org.}, 2016.
\newblock Accessed: 2019-11-15.

\bibitem{funke2018large}
Jan Funke, Fabian~David Tschopp, William Grisaitis, Arlo Sheridan, Chandan
  Singh, Stephan Saalfeld, and Srinivas~C Turaga.
\newblock Large scale image segmentation with structured loss based deep
  learning for connectome reconstruction.
\newblock {\em IEEE transactions on pattern analysis and machine intelligence},
  2018.

\bibitem{Gao_2019_ICCV}
Naiyu Gao, Yanhu Shan, Yupei Wang, Xin Zhao, Yinan Yu, Ming Yang, and Kaiqi
  Huang.
\newblock {SSAP}: Single-shot instance segmentation with affinity pyramid.
\newblock In {\em The IEEE International Conference on Computer Vision (ICCV)},
  October 2019.

\bibitem{grotschel1989cutting}
Martin Gr{\"o}tschel and Yoshiko Wakabayashi.
\newblock A cutting plane algorithm for a clustering problem.
\newblock {\em Mathematical Programming}, 45(1-3):59--96, 1989.

\bibitem{grotschel1990facets}
Martin Gr{\"o}tschel and Yoshiko Wakabayashi.
\newblock Facets of the clique partitioning polytope.
\newblock {\em Mathematical Programming}, 47(1-3):367--387, 1990.

\bibitem{januszewski2018high}
Micha{\l} Januszewski, J{\"o}rgen Kornfeld, Peter~H Li, Art Pope, Tim Blakely,
  Larry Lindsey, Jeremy Maitin-Shepard, Mike Tyka, Winfried Denk, and Viren
  Jain.
\newblock High-precision automated reconstruction of neurons with flood-filling
  networks.
\newblock {\em Nature methods}, 15(8):605, 2018.

\bibitem{johnson1967hierarchical}
Stephen~C Johnson.
\newblock Hierarchical clustering schemes.
\newblock {\em Psychometrika}, 32(3):241--254, 1967.

\bibitem{kappes2013comparative}
Joerg Kappes, Bjoern Andres, Fred Hamprecht, Christoph Schnorr, Sebastian
  Nowozin, Dhruv Batra, Sungwoong Kim, Bernhard Kausler, Jan Lellmann, Nikos
  Komodakis, et~al.
\newblock A comparative study of modern inference techniques for discrete
  energy minimization problems.
\newblock In {\em Proceedings of the IEEE conference on computer vision and
  pattern recognition}, pages 1328--1335, 2013.

\bibitem{kappes2011globally}
J{\"o}rg~Hendrik Kappes, Markus Speth, Bj{\"o}rn Andres, Gerhard Reinelt, and
  Christoph Schn{\"o}rr.
\newblock Globally optimal image partitioning by multicuts.
\newblock In {\em International Workshop on Energy Minimization Methods in
  Computer Vision and Pattern Recognition}, pages 31--44. Springer, 2011.

\bibitem{kardoostsolving}
Amirhossein Kardoost and Margret Keuper.
\newblock Solving minimum cost lifted multicut problems by node agglomeration.
\newblock In {\em ACCV 2018, 14th Asian Conference on Computer Vision}, Perth,
  Australia, 2018.

\bibitem{kaynig2015large}
Verena Kaynig, Amelio Vazquez-Reina, Seymour Knowles-Barley, Mike Roberts,
  Thouis~R Jones, Narayanan Kasthuri, Eric Miller, Jeff Lichtman, and Hanspeter
  Pfister.
\newblock Large-scale automatic reconstruction of neuronal processes from
  electron microscopy images.
\newblock {\em Medical image analysis}, 22(1):77--88, 2015.

\bibitem{kernighan1970efficient}
Brian~W Kernighan and Shen Lin.
\newblock An efficient heuristic procedure for partitioning graphs.
\newblock {\em Bell system technical journal}, 49(2):291--307, 1970.

\bibitem{keuper2015efficient}
Margret Keuper, Evgeny Levinkov, Nicolas Bonneel, Guillaume Lavou{\'e}, Thomas
  Brox, and Bjorn Andres.
\newblock Efficient decomposition of image and mesh graphs by lifted multicuts.
\newblock In {\em Proceedings of the IEEE International Conference on Computer
  Vision}, pages 1751--1759, 2015.

\bibitem{kiran2014global}
B~Ravi Kiran and Jean Serra.
\newblock Global--local optimizations by hierarchical cuts and climbing
  energies.
\newblock {\em Pattern Recognition}, 47(1):12--24, 2014.

\bibitem{kirillov2017instancecut}
Alexander Kirillov, Evgeny Levinkov, Bjoern Andres, Bogdan Savchynskyy, and
  Carsten Rother.
\newblock Instancecut: from edges to instances with multicut.
\newblock In {\em Proceedings of the IEEE Conference on Computer Vision and
  Pattern Recognition}, pages 5008--5017, 2017.

\bibitem{knowles2016rhoananet}
Seymour Knowles-Barley, Verena Kaynig, Thouis~Ray Jones, Alyssa Wilson, Joshua
  Morgan, Dongil Lee, Daniel Berger, Narayanan Kasthuri, Jeff~W Lichtman, and
  Hanspeter Pfister.
\newblock {RhoanaNet} pipeline: Dense automatic neural annotation.
\newblock {\em arXiv preprint arXiv:1611.06973}, 2016.

\bibitem{kokkinos2015pushing}
Iasonas Kokkinos.
\newblock Pushing the boundaries of boundary detection using deep learning.
\newblock {\em arXiv preprint arXiv:1511.07386}, 2015.

\bibitem{kong2018recurrentPix}
Shu Kong and Charless~C Fowlkes.
\newblock Recurrent pixel embedding for instance grouping.
\newblock In {\em Proceedings of the IEEE Conference on Computer Vision and
  Pattern Recognition}, pages 9018--9028, 2018.

\bibitem{krasowski2015improving}
Nikola Krasowski, Thorsten Beier, Graham~W Knott, Ullrich Koethe, Fred~A
  Hamprecht, and Anna Kreshuk.
\newblock Improving {3D} {EM} data segmentation by joint optimization over
  boundary evidence and biological priors.
\newblock In {\em 2015 IEEE 12th International Symposium on Biomedical Imaging
  (ISBI)}, pages 536--539. IEEE, 2015.

\bibitem{kunegis2010spectral}
J{\'e}r{\^o}me Kunegis, Stephan Schmidt, Andreas Lommatzsch, J{\"u}rgen Lerner,
  Ernesto~W De~Luca, and Sahin Albayrak.
\newblock Spectral analysis of signed graphs for clustering, prediction and
  visualization.
\newblock SIAM, 2010.

\bibitem{lange2018combinatorial}
Jan-Hendrik Lange, Bjoern Andres, and Paul Swoboda.
\newblock Combinatorial persistency criteria for multicut and max-cut.
\newblock In {\em Proceedings of the IEEE Conference on Computer Vision and
  Pattern Recognition}, pages 6093--6102, 2019.

\bibitem{lange2018partial}
Jan-Hendrik Lange, Andreas Karrenbauer, and Bjoern Andres.
\newblock Partial optimality and fast lower bounds for weighted correlation
  clustering.
\newblock In {\em International Conference on Machine Learning}, pages
  2898--2907, 2018.

\bibitem{lee2019learning}
Kisuk Lee, Ran Lu, Kyle Luther, and H~Sebastian Seung.
\newblock Learning dense voxel embeddings for {3D} neuron reconstruction.
\newblock {\em arXiv preprint arXiv:1909.09872}, 2019.

\bibitem{lee2017superhuman}
Kisuk Lee, Jonathan Zung, Peter Li, Viren Jain, and H~Sebastian Seung.
\newblock Superhuman accuracy on the {SNEMI3D} connectomics challenge.
\newblock {\em arXiv preprint arXiv:1706.00120}, 2017.

\bibitem{levinkov2017comparative}
Evgeny Levinkov, Alexander Kirillov, and Bjoern Andres.
\newblock A comparative study of local search algorithms for correlation
  clustering.
\newblock In {\em German Conference on Pattern Recognition}, pages 103--114.
  Springer, 2017.

\bibitem{liu2014modular}
Ting Liu, Cory Jones, Mojtaba Seyedhosseini, and Tolga Tasdizen.
\newblock A modular hierarchical approach to {3D} electron microscopy image
  segmentation.
\newblock {\em Journal of neuroscience methods}, 226:88--102, 2014.

\bibitem{liu2016image}
Ting Liu, Mojtaba Seyedhosseini, and Tolga Tasdizen.
\newblock Image segmentation using hierarchical merge tree.
\newblock {\em IEEE transactions on image processing}, 25(10):4596--4607, 2016.

\bibitem{liu2016sshmt}
Ting Liu, Miaomiao Zhang, Mehran Javanmardi, Nisha Ramesh, and Tolga Tasdizen.
\newblock {SSHMT}: Semi-supervised hierarchical merge tree for electron
  microscopy image segmentation.
\newblock In {\em European Conference on Computer Vision}, pages 144--159.
  Springer, 2016.

\bibitem{liu2018affinity}
Yiding Liu, Siyu Yang, Bin Li, Wengang Zhou, Jizheng Xu, Houqiang Li, and Yan
  Lu.
\newblock Affinity derivation and graph merge for instance segmentation.
\newblock In {\em Proceedings of the European Conference on Computer Vision
  (ECCV)}, pages 686--703, 2018.

\bibitem{malmberg2011generalized}
Filip Malmberg, Robin Strand, and Ingela Nystr{\"o}m.
\newblock Generalized hard constraints for graph segmentation.
\newblock In {\em Scandinavian Conference on Image Analysis}, pages 36--47.
  Springer, 2011.

\bibitem{meirovitch2016multi}
Yaron Meirovitch, Alexander Matveev, Hayk Saribekyan, David Budden, David
  Rolnick, Gergely Odor, Seymour Knowles-Barley, Thouis~Raymond Jones,
  Hanspeter Pfister, Jeff~William Lichtman, et~al.
\newblock A multi-pass approach to large-scale connectomics.
\newblock {\em arXiv preprint arXiv:1612.02120}, 2016.

\bibitem{milligan1979ultrametric}
Glenn~W Milligan.
\newblock Ultrametric hierarchical clustering algorithms.
\newblock {\em Psychometrika}, 44(3):343--346, 1979.

\bibitem{neven2019instance}
Davy Neven, Bert~De Brabandere, Marc Proesmans, and Luc~Van Gool.
\newblock Instance segmentation by jointly optimizing spatial embeddings and
  clustering bandwidth.
\newblock In {\em Proceedings of the IEEE Conference on Computer Vision and
  Pattern Recognition}, pages 8837--8845, 2019.

\bibitem{newell2017associative}
Alejandro Newell, Zhiao Huang, and Jia Deng.
\newblock Associative embedding: End-to-end learning for joint detection and
  grouping.
\newblock In {\em Advances in Neural Information Processing Systems}, pages
  2277--2287, 2017.

\bibitem{nunez2013machine}
Juan Nunez-Iglesias, Ryan Kennedy, Toufiq Parag, Jianbo Shi, and Dmitri~B
  Chklovskii.
\newblock Machine learning of hierarchical clustering to segment {2D} and {3D}
  images.
\newblock {\em PloS one}, 8(8):e71715, 2013.

\bibitem{pape2017solving}
Constantin Pape, Thorsten Beier, Peter Li, Viren Jain, Davi~D Bock, and Anna
  Kreshuk.
\newblock Solving large multicut problems for connectomics via domain
  decomposition.
\newblock In {\em Proceedings of the IEEE International Conference on Computer
  Vision}, pages 1--10, 2017.

\bibitem{perlin1985image}
Ken Perlin.
\newblock An image synthesizer.
\newblock {\em ACM Siggraph Computer Graphics}, 19(3):287--296, 1985.

\bibitem{perlin2001noise}
Ken Perlin.
\newblock Noise hardware.
\newblock {\em Real-Time Shading SIGGRAPH Course Notes}, 2001.

\bibitem{rangapuram2012constrained}
Syama~Sundar Rangapuram and Matthias Hein.
\newblock Constrained 1-spectral clustering.
\newblock In {\em AISTATS}, volume~30, page~90, 2012.

\bibitem{ren2013image}
Zhile Ren and Gregory Shakhnarovich.
\newblock Image segmentation by cascaded region agglomeration.
\newblock In {\em Proceedings of the IEEE Conference on Computer Vision and
  Pattern Recognition}, pages 2011--2018, 2013.

\bibitem{ronneberger2015u}
Olaf Ronneberger, Philipp Fischer, and Thomas Brox.
\newblock U-net: Convolutional networks for biomedical image segmentation.
\newblock In {\em International Conference on Medical image computing and
  computer-assisted intervention}, pages 234--241. Springer, 2015.

\bibitem{salembier2000binary}
Philippe Salembier and Luis Garrido.
\newblock Binary partition tree as an efficient representation for image
  processing, segmentation, and information retrieval.
\newblock {\em IEEE transactions on Image Processing}, 9(4):561--576, 2000.

\bibitem{schmidt2018cell}
Uwe Schmidt, Martin Weigert, Coleman Broaddus, and Gene Myers.
\newblock Cell detection with star-convex polygons.
\newblock In {\em International Conference on Medical Image Computing and
  Computer-Assisted Intervention}, pages 265--273. Springer, 2018.

\bibitem{sofiiuk2019adaptis}
Konstantin Sofiiuk, Olga Barinova, and Anton Konushin.
\newblock {AdaptIS}: Adaptive instance selection network.
\newblock In {\em Proceedings of the IEEE International Conference on Computer
  Vision}, pages 7355--7363, 2019.

\bibitem{turaga2009maximin}
Srinivas~C Turaga, Kevin~L Briggman, Moritz Helmstaedter, Winfried Denk, and
  H~Sebastian Seung.
\newblock Maximin affinity learning of image segmentation.
\newblock pages 1865--1873, 2009.

\bibitem{uzunbas2016efficient}
Mustafa~Gokhan Uzunbas, Chao Chen, and Dimitris Metaxas.
\newblock An efficient conditional random field approach for automatic and
  interactive neuron segmentation.
\newblock {\em Medical image analysis}, 27:31--44, 2016.

\bibitem{wang2014constrained}
Xiang Wang, Buyue Qian, and Ian Davidson.
\newblock On constrained spectral clustering and its applications.
\newblock {\em Data Mining and Knowledge Discovery}, 28(1):1--30, 2014.

\bibitem{wolf2019mutex}
Steffen Wolf, Alberto Bailoni, Constantin Pape, Nasim Rahaman, Anna Kreshuk,
  Ullrich K{\"o}the, and Fred~A Hamprecht.
\newblock The mutex watershed and its objective: Efficient, parameter-free
  graph partitioning.
\newblock {\em IEEE transactions on pattern analysis and machine intelligence},
  43(10):3724--3738, 2020.

\bibitem{wolf2018mutex}
Steffen Wolf, Constantin Pape, Alberto Bailoni, Nasim Rahaman, Anna Kreshuk,
  Ullrich Kothe, and FredA Hamprecht.
\newblock The mutex watershed: Efficient, parameter-free image partitioning.
\newblock In {\em Proceedings of the European Conference on Computer Vision
  (ECCV)}, pages 546--562, 2018.

\bibitem{xie2015holistically}
Saining Xie and Zhuowen Tu.
\newblock Holistically-nested edge detection.
\newblock In {\em Proc. ICCV'15}, pages 1395--1403, 2015.

\bibitem{yarkony2012fast}
Julian Yarkony, Alexander Ihler, and Charless~C Fowlkes.
\newblock Fast planar correlation clustering for image segmentation.
\newblock In {\em European Conference on Computer Vision}, pages 568--581.
  Springer, 2012.

\bibitem{zeng2017deepem3d}
Tao Zeng, Bian Wu, and Shuiwang Ji.
\newblock {DeepEM3D}: approaching human-level performance on {3D} anisotropic
  {EM} image segmentation.
\newblock {\em Bioinformatics}, 33(16):2555--2562, 2017.

\end{thebibliography}
